\documentclass[final,3p,times]{elsarticle}

\usepackage{amssymb}
\usepackage{lipsum}
\usepackage{graphicx}
\usepackage{color}
\usepackage{bm}
\usepackage{amsmath}
\usepackage{slashed}
\usepackage{mathrsfs}
\usepackage[normalem]{ulem}
\usepackage{comment}

\usepackage{dcolumn}
\usepackage{hyperref}
\hypersetup{colorlinks=true, citecolor=blue, urlcolor=blue, linkcolor=blue}
\usepackage{epstopdf}
\usepackage{subcaption}

\newcommand{\dd}{\mathrm{d}}

\journal{Physica A: Statistical Mechanics and its Applications}

\begin{document}

\begin{frontmatter}

\title{Posterior Collapse as a Phase Transition in Variational Autoencoders}

\author[first]{Zhen Li}
\ead{li-zhen@g.ecc.u-tokyo.ac.jp}
\affiliation[first]{organization={Department of Complexity Science and Engineering, Graduate School of Frontier Sciences, The University of Tokyo},
            city={Kashiwa},
            postcode={277-8561}, 
            state={Chiba},
            country={Japan}}

\author[second]{Fan Zhang}
\ead{zhang-fan@g.ecc.u-tokyo.ac.jp}
\affiliation[second]{organization={Department of Human and Engineered Environmental Studies, Graduate School of Frontier Sciences, The University of Tokyo},
            city={Kashiwa},
            postcode={277-8561}, 
            state={Chiba},
            country={Japan}}

\author[second]{Zheng Zhang}
\ead{mark-zhang@g.ecc.u-tokyo.ac.jp}

\author[second]{Yu Chen}
\ead{chen@edu.k.u-tokyo.ac.jp}

\begin{abstract}
    We investigate the phenomenon of posterior collapse in variational autoencoders (VAEs) from the perspective of statistical physics, and reveal that it constitutes a phase transition governed jointly by data structure and model hyper-parameters.
    By analyzing the stability of the trivial solution associated with posterior collapse, we identify a critical hyper-parameter threshold.
    In particular, we derive an explicit criterion for the onset of collapse: posterior collapse occurs when the decoder variance exceeds the largest eigenvalue of the data covariance matrix.
    This critical boundary, separating meaningful latent inference from collapse, is characterized by a discontinuity in the KL divergence between the approximate posterior and the prior distribution, where the KL divergence and its derivatives exhibit clear non-analytic behavior.
    We validate this critical behavior on both synthetic and real-world datasets, confirming the existence of a phase transition.
    The experimental results align well with our theoretical predictions, demonstrating the robustness of our collapse criterion across various VAE architectures.
    Our stability-based analysis demonstrate that posterior collapse is not merely an optimization failure, but rather an emerging phase transition arising from the interplay between data structure and variational constraints.
    This perspective offers new insights into the trainability and representational capacity of deep generative models.
\end{abstract}

\begin{keyword}
Variational autoencoders \sep Posterior collapse \sep Phase transition
\end{keyword}

\end{frontmatter}

\section{Introduction}

Variational autoencoders (VAEs) are a fundamental class of latent-variable generative models that combine variational inference with deep neural networks~\cite{kingma2013auto, girin2020dynamical, blei2017variational}. 
By introducing a tractable variational posterior, VAEs enable approximate Bayesian learning in high-dimensional spaces, and their training objective, evidence lower bound (ELBO), balances reconstruction fidelity with a KL-regularization term that controls the complexity of the latent representation~\cite{kingma2013auto, girin2020dynamical, blei2017variational, jordan1999introduction}. 
This framework has led to widespread success in applications including image synthesis~\cite{huang2018introvae}, anomaly detection~\cite{lin2020anomaly}, and molecular design~\cite{dollar2023efficient}.

Despite this success, VAEs frequently exhibit \emph{posterior collapse}---a phenomenon in which latent variables become uninformative and the approximate posterior degenerates to the prior~\cite{lucas2019don, dang2023beyond, wang2022posterior, dai2020usual}. 
Posterior collapse poses a major limitation for representation learning, especially when the decoder is expressive or the dataset exhibits strong correlations.
Classical explanations attribute collapse to the KL term in the ELBO, which encourages the posterior distribution to match the prior~\cite{bowman2015generating, kingma2016improved, sonderby2016ladder, huang2018improving}. 
Numerous heuristics such as $\beta$-VAE~\cite{higgins2017beta}, KL annealing~\cite{huang2018improving}, or modified ELBO formulations~\cite{hoffman2016elbo} have been proposed, yet a principled understanding of \emph{when} collapse occurs remains incomplete.

In parallel to these theoretical analyses, several recent works have explored improving the structure or expressiveness of the latent space in practical VAE applications. 
Dimension-weighting methods~\cite{liu2024improving} enhance disentanglement by exploiting feature imbalance; Cloud-VAE~\cite{liu2023cloud} introduces concept-level structure into the latent representation; and PixelVAE~\cite{gulrajani2016pixelvae} combines VAEs with autoregressive PixelCNN decoders to improve sample quality. 
These approaches focus on enhancing the semantic or expressive properties of the latent space, which is complementary to our goal of characterizing the conditions under which the latent variables become entirely uninformative due to posterior collapse.

Recent analytical studies have provided deeper insight by examining \emph{linear VAEs}~\cite{lucas2019don, dang2023beyond, wang2022posterior, dai2020usual}. 
These works reveal roles played by the log marginal likelihood~\cite{lucas2019don}, posterior geometry~\cite{dang2023beyond}, competition among ELBO terms~\cite{wang2022posterior}, and local extrema~\cite{dai2020usual}. 
More recently, Ichikawa and Hukushima~\cite{ichikawa2025high} analyzed a linear $\beta$-VAE using the replica method in a high-dimensional setting, deriving a phase diagram with respect to the regularization strength in ELBO and sample complexity. 
While these results offer valuable asymptotic characterizations, they fundamentally rely on strict linear encoder-decoder mappings and do not provide an architecture-independent condition applicable to practical non-linear VAEs.

In light of these limitations, there remains a need for a general theoretical framework that characterizes the onset of posterior collapse \emph{beyond} the linear setting, and that directly connects collapse to statistical properties of the data and model hyperparameters. 
Our objective in this work is to provide such a framework by interpreting posterior collapse as a \emph{phase transition} in the ELBO landscape.

In this paper, we make the following contributions:
\begin{enumerate}
    \item We reinterpret posterior collapse as the gain of stability of the trivial ELBO extremum, providing a statistical-physics perspective on collapse in VAEs.
    \item We derive a general and architecture-independent criterion for the onset of collapse.  
    Specifically, for deep Gaussian VAEs, collapse occurs when the decoder variance exceeds the largest eigenvalue of the data covariance matrix.
    \item We show that the KL divergence exhibits non-analytic behavior at the predicted critical point, consistent with phase-transition theory.
    \item We validate the theoretical prediction using both synthetic Gaussian data and real-world datasets, demonstrating that the criterion accurately matches the empirical collapse boundary in non-linear VAEs.
\end{enumerate}

The rest of this paper is organized as follows.
Section~\ref{sec_rev} reviews the ELBO formulation of VAEs and characterizes the corresponding extrema.
Section~\ref{sec_post} presents the collapse criterion based on the analysis of stability, interpretes posterior collapse as a phase transition, and provides validation on synthetic and real datasets.
A broader discussion of the practical implications of our findings is provided in Section~\ref{sec_applicability}.
Section~\ref{sec_con} concludes with discussion and future directions.

\section{Related works}
\textbf{Bayesian deep learning}: VAEs represent a central approach in Bayesian deep learning, combining deep neural networks with variational inference to approximate intractable posteriors~\cite{kingma2013auto, blei2017variational}.
The ELBO objective provides a principled way to learn latent-variable generative models by balancing reconstruction fidelity and latent regularization.
Subsequent developments such as $\beta$-VAE~\cite{higgins2017beta} and ladder VAE~\cite{sonderby2016ladder} further explore trade-offs between disentanglement, expressivity, and inference accuracy.
Recent surveys~\cite{girin2020dynamical} have highlighted the growing diversity of Bayesian generative models, including extensions to hierarchical and dynamic settings.

\textbf{Posterior collapse}: Despite their flexibility, VAEs are known to suffer from posterior collapse, where the approximate posterior becomes indistinguishable from the prior, rendering latent variables uninformative~\cite{bowman2015generating,lucas2019don,dai2020usual}.
Early explanations attributed this issue to the strong regularizing effect of the KL-divergence in the ELBO, motivating strategies such as KL annealing and $\beta$-VAE modifications~\cite{huang2018improving,higgins2017beta}.
Later work examined additional causes, including poor decoder architecture, inference mismatch, and suboptimal optimization dynamics~\cite{cremer2018inference, hoffman2016elbo}.
More recently, researchers have proposed that posterior collapse can result from deeper statistical properties of the learning objective itself~\cite{lucas2019don,dang2023beyond,wang2022posterior}, hinting at the need for a more principled theoretical framework.

\textbf{Linear models and analytical perspectives}: To better understand the mechanism of posterior collapse, several works have studied VAEs in linear settings where analytical insights are tractable~\cite{lucas2019don,dang2023beyond,wang2022posterior,ichikawa2025high}.
These studies reveal that collapse can occur even in the absence of optimization failures, due to properties intrinsic to the ELBO landscape—such as the dominance of the log marginal likelihood or the flatness of local extrema.
These findings have motivated our approach, which extends beyond linear assumptions by analyzing posterior collapse as a phase transition determined jointly by data structure and model hyper-parameters.

Beyond the theoretical analyses of posterior collapse, a parallel line of recent VAE research  focuses on improving the structure and interpretability of the latent space. 
Dimension-weighting methods~\cite{liu2024improving} exploit feature imbalance in VAEs to enhance disentanglement by reweighting latent dimensions, while Cloud-VAE~\cite{liu2023cloud} introduces concept embeddings to impose interpretable semantic structure in the latent space. 
Other approaches such as PixelVAE~\cite{gulrajani2016pixelvae} combine autoregressive PixelCNN decoders with VAEs to improve expressiveness in modeling natural images.
These works aim to enhance the quality, semantics, or expressiveness of latent representations. 
In contrast, the present study examines a different question: 
\emph{under what conditions does the latent representation become completely uninformative?} 

While prior studies have significantly advanced the understanding of posterior collapse, most analyses rely on specific structural assumptions that limit their generality. 
For instance, analytical treatments based on linear VAEs~\cite{lucas2019don, wang2022posterior, dai2020usual} offer valuable intuition but do not capture the behavior of practical non-linear models. 
Replica-method analyses~\cite{ichikawa2025high} provide elegant asymptotic characterizations, yet fundamentally depend on strict linear encoder-decoder mappings and a particular high-dimensional limit. 
Other explanations focusing on ELBO competition or posterior mismatch~\cite{dang2023beyond} illuminate contributing factors but do not yield an explicit condition predicting when collapse occurs.

These limitations highlight a gap in the existing literature: there is no architecture-independent, theoretically grounded criterion that determines the onset of posterior collapse in general deep VAEs. 
Our work addresses this gap by analyzing the stability of the ELBO's trivial solution and deriving a collapse condition that depends solely on the data covariance spectrum and decoder variance, independent of model architecture or linearity assumptions.

\section{Maximizing evidence lower bound}
\label{sec_rev}
A VAE compresses high-dimensional data $\bm{x} \in \mathbb{R}^N$ into low-dimensional latent variables $\bm{z} \in \mathbb{R}^n$ (with $N > n$) using a prior distribution $p_{\rm lat}(\bm{z})$  via an encoder $q_{\phi}(\bm{z}|\bm{x})$, and generate data from latent variables via a decoder $p_{\theta}(\bm{x}|\bm{z})$~\cite{kingma2013auto, girin2020dynamical}.
Due to the fact that the distribution $p_{\rm data}(\bm{x})$ for data $\bm{x}$ is usually intractable, the encoder $q_{\phi}(\bm{z}|\bm{x})$, also known as variational posterior distribution, approximates the posterior distribution $p(\bm{z}|\bm{x})$ via amortized variational inference~\cite{dang2023beyond, blei2017variational}, which is tractable using the evidence lower bound (ELBO)~\cite{kingma2013auto}:
\begin{align}
    {\rm ELBO}(\bm{x}) &\equiv \mathbb{E}_{q_{\phi}(\bm{z}|\bm{x})}\left[\ln p_{\theta}(\bm{x}|\bm{z})\right] - D_{\rm KL}\left(q_{\phi}(\bm{z}|\bm{x}) || p_{\rm lat}(\bm{z})\right)\leq \ln p_{\rm data}(\bm{x}).
\end{align}
Training a VAE aims to maximize the ELBO, which includes the reconstruction loss $\mathbb{E}_{q_{\phi}(\bm{z}|\bm{x})}\left[\ln p_{\theta}(\bm{x}|\bm{z})\right]$ and the KL-divergence $D_{\rm KL}\left(q_{\phi}(\bm{z}|\bm{x}) || p_{\rm lat}(\bm{z})\right)$ as regularization~\cite{kingma2013auto, dang2023beyond}.
As large amount of data $\bm{x}$ are used for training, the target for a VAE can be also regarded as the expectation of ELBO~\cite{rezende2018taming,wang2022posterior}:

\begin{align}
    \mathcal{L}_{\rm VAE} &\equiv \mathbb{E}_{p_{\rm data}(\bm{x})}\left[{\rm ELBO}(\bm{x})\right]\notag\\
    &= \mathbb{E}_{p_{\rm data}(\bm{x})}\left[\mathbb{E}_{\bm{z}\sim q_{\phi}(\bm{z}|\bm{x})}\left[\ln p_{\theta}(\bm{x}|\bm{z})\right]\right]-\mathbb{E}_{p_{\rm data}(\bm{x})}\left[D_{\rm KL}\left(q_{\phi}(\bm{z}|\bm{x}) || p_{\rm lat}(\bm{z})\right)\right]\notag\\
    &= \int \ln p_{\theta}(\bm{x}|\bm{z})q_{\phi}(\bm{z}|\bm{x})p_{\rm data}(\bm{x}){\rm d}\bm{x}{\rm d}\bm{z}-\int q_{\phi}(\bm{z}|\bm{x})\ln\frac{q_{\phi}(\bm{z}|\bm{x})}{p_{\rm lat}(\bm{z})}p_{\rm data}(\bm{x}){\rm d}\bm{x}{\rm d}\bm{z}.\label{eq:elbo}
\end{align}

The extrema of $\mathcal{L}_{\rm VAE}$ can be obtained by performing variational calculus on
\begin{align}
    \mathcal{I}[p_{\theta}(\bm{x}|\bm{z}),q_{\phi}(\bm{z}|\bm{x})] &\equiv \mathcal{L}_{\rm VAE} - \lambda_1(\bm{z})\left[\int p_{\theta}(\bm{x}|\bm{z}){\rm d}\bm{x} -1\right] - \lambda_2(\bm{x})\left[\int q_{\phi}(\bm{z}|\bm{x}){\rm d}\bm{z} -1\right],\label{eq:vae_constraint}
\end{align}
where Lagrange multipliers $\lambda_1(\bm{z})$ and $\lambda_2(\bm{x})$ are introduced for the normalizations
\begin{align}
    \int p_{\theta}(\bm{x}|\bm{z}){\rm d}\bm{x} &=1,\\
    \int q_{\phi}(\bm{z}|\bm{x}){\rm d}\bm{z} &= 1,
\end{align}
respectively.
The extrema points reads:
\begin{align}
    p_{\theta}(\bm{x}|\bm{z}) &= p(\bm{x}|\bm{z})\label{eq_vae_pxz}\\
    q_{\phi}(\bm{z}|\bm{x}) &= p(\bm{z}|\bm{x}),\label{eq_vae_pzx}
\end{align}
where the multipliers are
\begin{align}
    \lambda_1(\bm{z}) &= p_{\rm lat}(\bm{z})\\
    \lambda_2(\bm{x}) &= p_{\rm data}(\bm{x})\left[\ln p_{\rm data}(\bm{x}) -1\right].
\end{align}
The meaning of the extrema is intuitive: the conditional distributions $p(\bm{x}|\bm{z})$ and $p(\bm{z}|\bm{x})$ are required to perfectly encode and reconstruct the data from the latent variables.
In this case, it is easy to check the extrema value
\begin{equation}
    \mathcal{L}_{\rm VAE}^* = \int p_{\rm data}(\bm{x})\ln p_{\rm data}(\bm{x})\dd \bm{x},\label{eq:vae_extrema}
\end{equation}
which is the log likelihood, or the opposite of the entropy, of data.

\begin{figure}[htbp]
    \centering
    \includegraphics[width=0.5\linewidth]{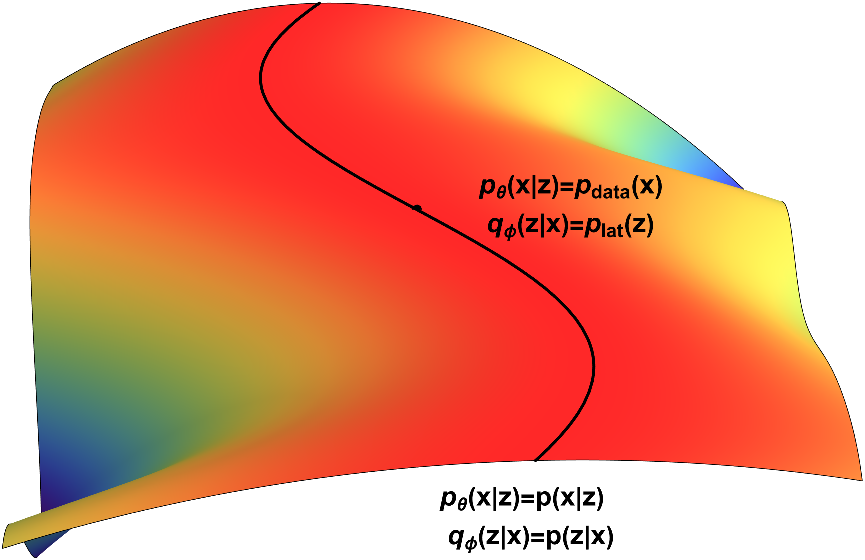}
    \captionsetup{justification=raggedright,singlelinecheck=false}
    \caption
    {
        Sketch of the $\mathcal{L}_{\rm VAE}$ surface. 
        There exists multiple extrema points (black curve), around which the surface is flat along certain directions. 
        Posterior collapse $p_{\theta}(\bm{x}|\bm{z}) = p_{\rm data}(\bm{x})$, $q_{\phi}(\bm{z}|\bm{x})= p_{\rm lat}(\bm{z})$ (black dot) is also an extrema point.
    }
    \label{fig:flatness}
\end{figure}

It is worth noting that there are multiple extrema points, as there is more than one set of conditional distributions.
The existence of multiple extrema of the ELBO has also been noted in previous works, such as~\cite{tipping1999probabilistic, kunin2019loss}.
We can also understand it by analyzing the stability of extrema points.
By applying second order variation on Eq.~\eqref{eq:vae_constraint} and substitute Eqs.~\eqref{eq_vae_pxz} and \eqref{eq_vae_pzx}, we have the Hessian matrix $\bm{H}(\bm{x}, \bm{y})$ for the extrema points as:
\begin{align}
    \bm{H}(\bm{x}, \bm{y}) &=\left.
    \begin{pmatrix}
        \dfrac{\delta^2 \mathcal{I}}{\delta p_{\theta}^2} & \dfrac{\delta^2 \mathcal{I}}{\delta p_{\theta} \delta q_\phi}\\
        \\
        \dfrac{\delta^2 \mathcal{I}}{\delta p_{\theta} \delta q_\phi} & \dfrac{\delta^2 \mathcal{I}}{\delta q_{\phi}^2}
    \end{pmatrix}\right|_{\substack{p_{\theta}(\bm{x}|\bm{z}) =p(\bm{x}|\bm{z}) \\ q_{\phi}(\bm{z}|\bm{x})= p(\bm{z}|\bm{x})}} = 
    \begin{pmatrix}
        -\dfrac{p(\bm{z}|\bm{x})}{p^2(\bm{x}|\bm{z})} & \dfrac{1}{p(\bm{x}|\bm{z})}\\
        \\
        \dfrac{1}{p(\bm{x}|\bm{z})} & -\dfrac{1}{q(\bm{z}|\bm{x})}
    \end{pmatrix}p_{\rm data}(\bm{x}),
\end{align}
which is negative semi-definite, implying the flatness along certain directions around extrema points.
This further implies the existence of multiple extrema, as illustrated in Figure \ref{fig:flatness}.

There is a special extrema point,
\begin{align}
    p_{\theta}(\bm{x}|\bm{z}) &= p_{\rm data}(\bm{x}),\\
    q_{\phi}(\bm{z}|\bm{x}) &= p_{\rm lat}(\bm{z}),
\end{align}
known as posterior collapse (also in Figure \ref{fig:flatness}), where the variational posterior distribution $q_{\phi}(\bm{z}|\bm{x})$ converges to the prior distribution $p_{\rm lat}(\bm{z})$, rendering the latent variables uninformative about the data,  and limiting the capacity of VAEs to extract and represent meaningful structure from the data~\cite{wang2022posterior, bowman2015generating, dang2023beyond, lucas2019don}.

In practice, directly approximating the true posterior distribution is challenging.
The standard approach is to assume specific forms for $p_{\theta}(\bm{x}|\bm{z})$ and $q_{\phi}(\bm{z}|\bm{x})$, and to approximate parameters such as the mean and variance using deep learning.
One of the commonly chosen form reads:
\begin{align}
    p_{\theta}(\bm{x}|\bm{z})&\equiv \prod_{i=1}^N \frac{1}{\sqrt{2\pi}\sigma'}\exp\left\{-\frac{[x_i - f_i(\bm{z})]^2}{2\sigma'^2}\right\},\label{eq:vae_gaussian_pxz}\\
    q_{\phi}(\bm{z}|\bm{x})&\equiv \prod_{j=1}^n \frac{1}{\sqrt{2\pi}\sigma_j(\bm{x})}\exp\left\{-\frac{[z_j - \mu_j(\bm{x})]^2}{2\sigma_j^2(\bm{x})}\right\},\label{eq:vae_gaussian_pzx}
\end{align}
while the prior distribution
\begin{equation}
    p_{\rm lat}(\bm{z}) \equiv \prod_{j=1}^n \frac{1}{\sqrt{2\pi}}\exp\left(-\frac{z_j^2}{2}\right),\label{eq:vae_gaussian_pz}
\end{equation}
is standard Gaussian.
This form is known as deep Gaussian VAEs, as both $p_{\theta}(\bm{x}|\bm{z})$ and $q_{\phi}(\bm{z}|\bm{x})$ are Gaussian with the parameters $\bm{f}(\bm{z})$, $\bm{\mu}(\bm{x})$ and $\bm{\sigma}(\bm{x})$ learned from networks~\cite{lucas2019don, rezende2018taming}.
The hyper-parameter $\sigma'^2$ describes the reconstruction noise variance.

Substitute Eqs.~\eqref{eq:vae_gaussian_pxz}--\eqref{eq:vae_gaussian_pz} into Eq.~\eqref{eq:elbo}, we have the training target for deep Gaussian VAEs~\cite{rezende2018taming,lucas2019don,wang2022posterior}:
\begin{align}
    \mathcal{L}_{\rm VAE}^{\rm G} &\equiv N\ln\frac{1}{\sqrt{2\pi}\sigma'}-\int \frac{1}{2\sigma'^2}\sum_{i=1}^{N}\left[x_i-f_i(\bm{z})\right]^2q_{\phi}(\bm{z}|\bm{x})p_{\rm data}(\bm{x}){\rm d}\bm{x}{\rm d}\bm{z}\notag\\ 
    &\qquad- \frac{1}{2}\int\sum_{j=1}^{n}\left[\sigma^2_j(\bm{x})+\mu_j^2(\bm{x})-1-\ln \sigma_j^2(\bm{x})\right]p_{\rm data}(\bm{x}){\rm d}\bm{x}.\label{eq:vae_gaussian_elbo}
\end{align}
By applying variation on Eq.~\eqref{eq:vae_gaussian_elbo}, we derive the conditions for extrema points:
\begin{align}
    f_i(\bm{z}) &= \frac{\displaystyle \int x_i q_{\phi}(\bm{z}|\bm{x})p_{\rm data}(\bm{x}){\rm d}\bm{x}}{\displaystyle \int q_{\phi}(\bm{z}|\bm{x})p_{\rm data}(\bm{x}){\rm d}\bm{x}},\\
    \mu_j(\bm{x}) &= \frac{\displaystyle \frac{1}{2\sigma'^2}\int \sum_{i=1}^{N}\left[x_i-f_i(\bm{z})\right]^2 z_j q_{\phi}(\bm{z}|\bm{x}){\rm d}\bm{z}}{\displaystyle \int \sum_{i=1}^{N}\left[x_i-f_i(\bm{z})\right]^2q_{\phi}(\bm{z}|\bm{x}){\rm d}\bm{z}-\sigma_j^2(\bm{x})},
\end{align}
and
\begin{align}
    &\sigma^4_j(\bm x) - \left\{1+\frac{1}{2\sigma'^2}\int \sum_{i=1}^{N}\left[x_i-f_i(\bm{z})\right]^2 q_{\phi}(\bm{z}|\bm{x}){\rm d}\bm{z}\right\}\sigma^2_j(\bm x)+ \frac{1}{2\sigma'^2}\int \sum_{i=1}^{N}\left[x_i-f_i(\bm{z})\right]^2 \left[z_j-\mu_j(\bm{z})\right]^2 q_{\phi}(\bm{z}|\bm{x}){\rm d}\bm{z} = 0.
\end{align}
It is easy to find a trivial extrema point:
\begin{align}
    f_i(\bm z) &= \left<x_i\right>,\label{eq:extrema_1}\\
    \mu_j(\bm x) &= 0,\\
    \sigma_j(\bm x) &= 1,\label{eq:extrema_3}
\end{align}
which corresponds to posterior collapse, as $q_{\phi}(\bm{z}|\bm{x}) = p_{\rm lat}(\bm z)$ at this point.
Here, we denote $\left<\theta\right> \equiv \mathbb{E}_{p_{\rm data}(\bm x)}[\theta]$.
We can also find the extrema value for posterior collapse,
\begin{equation}
    \mathcal{L}_{\rm VAE}^{\rm G*} = -\frac{N}{2}\ln 2\pi \sigma'^2 - \frac{1}{2\sigma'^2}\mathrm{Tr}(\bm{\Sigma}),\label{eq:vae_gaussian_extrema}
\end{equation}
where $\bm{\Sigma}$ is the covariance matrix for data $\bm{x}$.
It should be noted that this extremum, which also depends on the hyper-parameter $\sigma'^2$, generally does not coincide with the data's log-likelihood.

\section{Posterior collapse as phase transition}
\label{sec_post}
\begin{figure*}
    \captionsetup{justification=raggedright,singlelinecheck=true}
    \centering
    \begin{subfigure}[b]{0.32\textwidth}
        \centering
        \includegraphics[width=\linewidth]{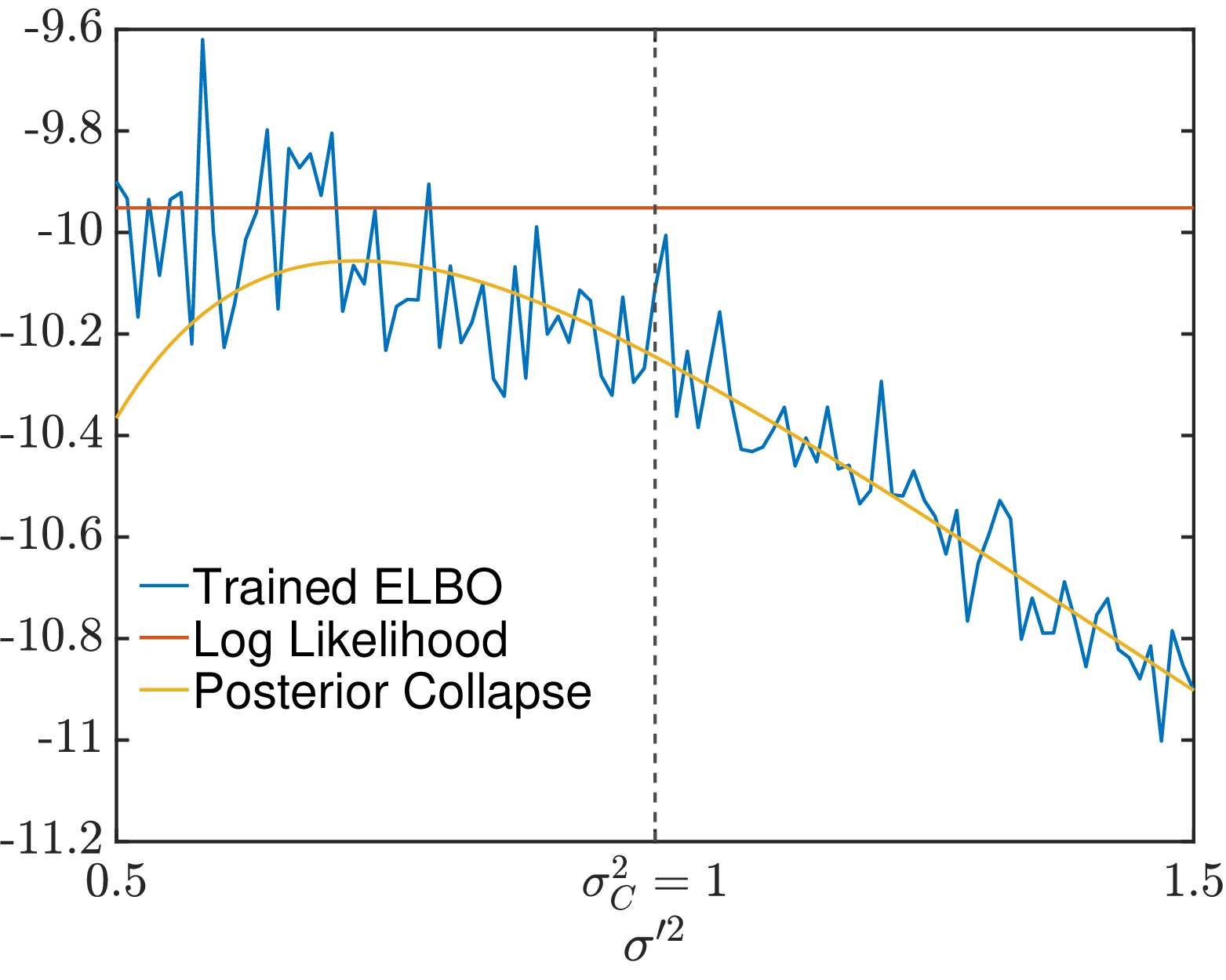}
        \caption{}
        \label{fig:elbo_n_4}
    \end{subfigure}
    \hfill
    \begin{subfigure}[b]{0.32\textwidth}
        \centering
        \includegraphics[width=\linewidth]{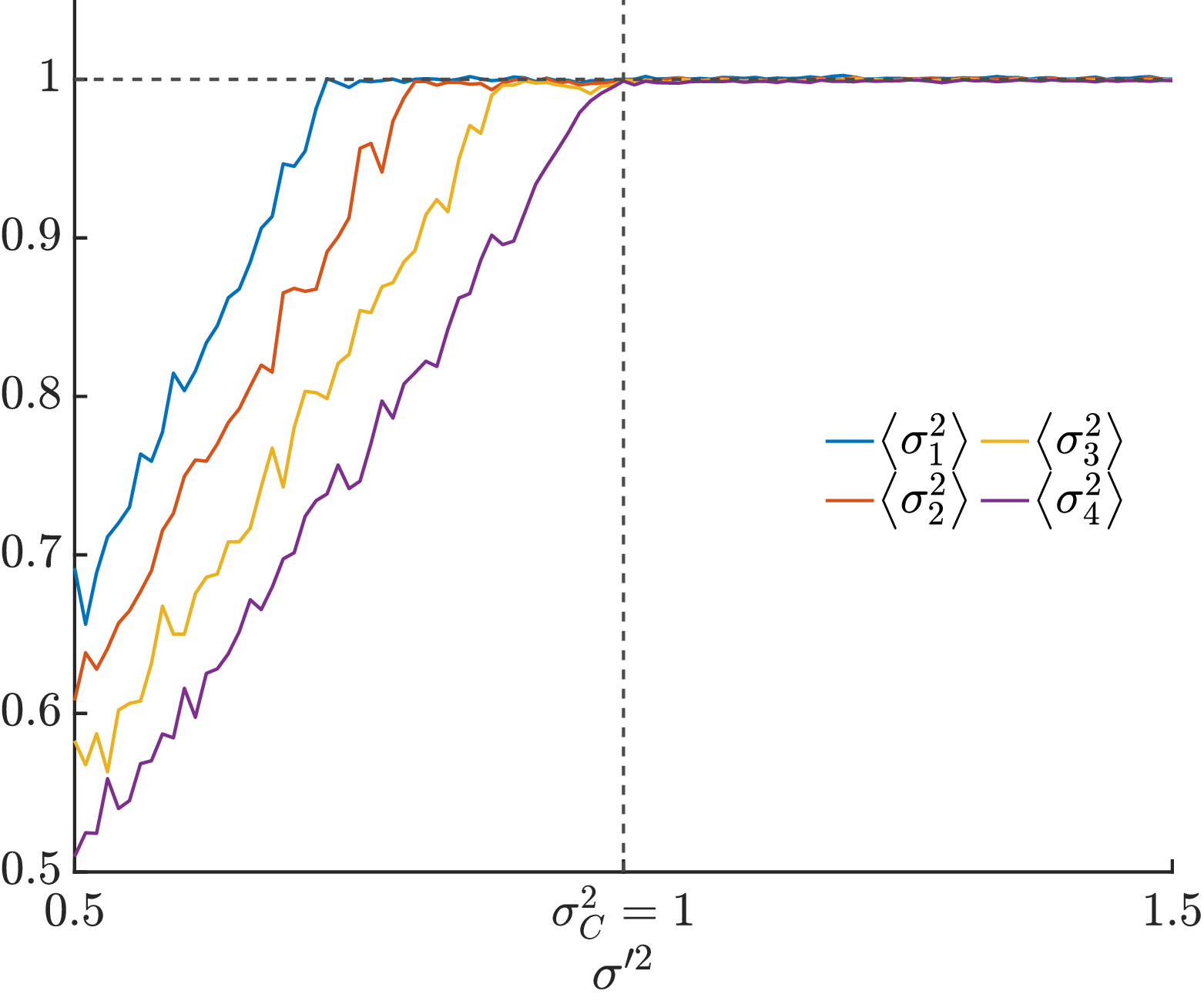}
        \caption{}
        \label{fig:var_n_4}
    \end{subfigure}
    \hfill
    \begin{subfigure}[b]{0.32\textwidth}
        \centering
        \includegraphics[width=\linewidth]{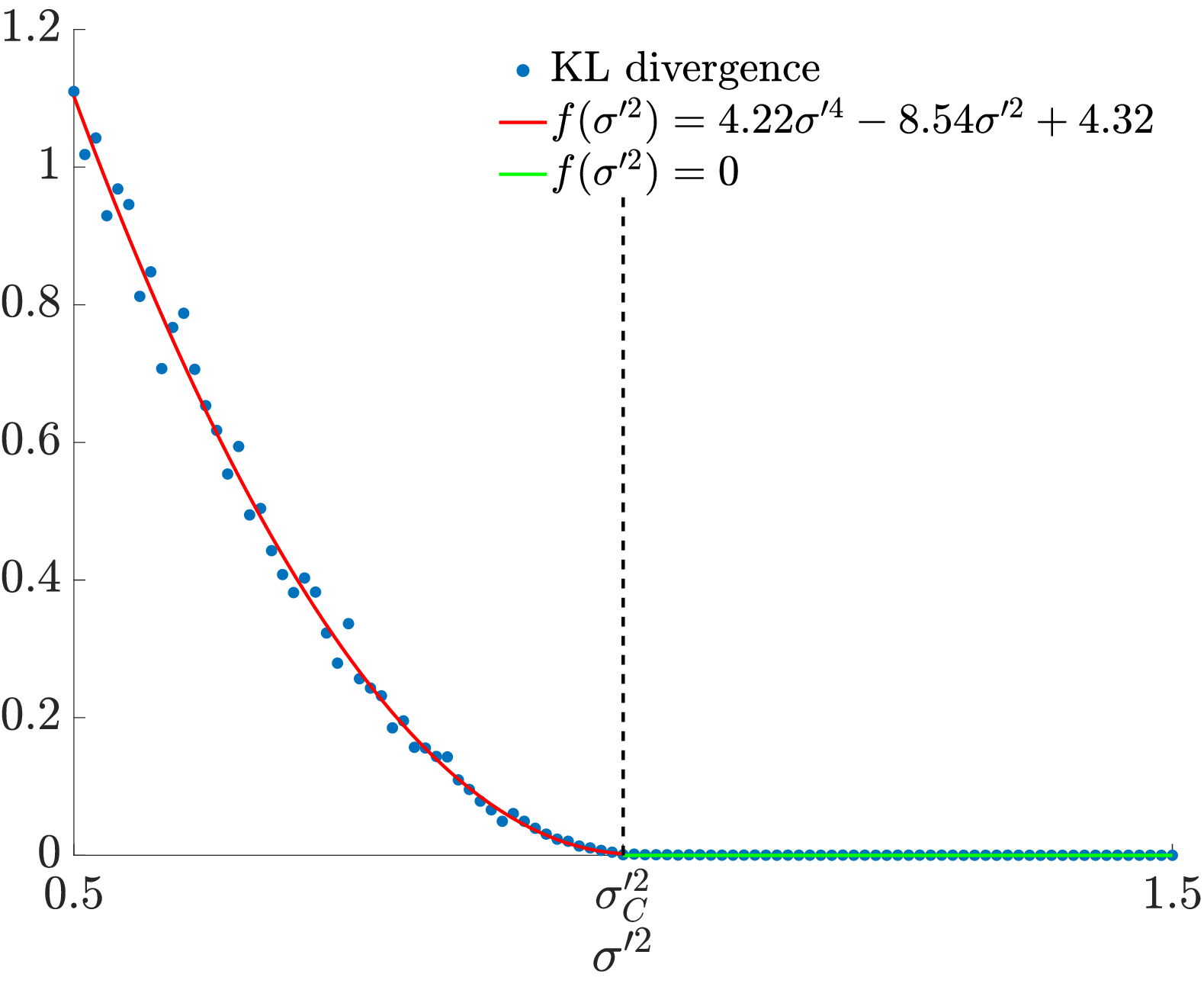}
        \caption{}
        \label{fig:kl_n_4}
    \end{subfigure}
    \caption{(a) Trained ELBO, log likelihood and posterior collapse line v.s. $\sigma'^2$. The trained ELBO converges the posterior collapse when $\sigma'^2>\sigma'^2_C =1$.
    (b) Mean latent variances $\left<\sigma_j^2\right>$ ($j=1,2,3,4$) v.s. $\sigma'^2$. All of them equal 1 when $\sigma'^2 >\sigma'^2_C=1$.
    (c) KL divergence v.s $\sigma'^2$, which converges to 0 when $\sigma'^2>\sigma'^2_C = 1$ (green line), and can be fitted by a quadratic function of $\sigma'^2$ when $\sigma'^2<\sigma'^2_C = 1$ (red line).
    The fitted function is used to illustrate the non-analytic derivative behavior near the critical point.}
\end{figure*}

\subsection*{Criterion}
Previous studies on posterior collapse have primarily focused on linear VAE models~\cite{lucas2019don, dang2023beyond, wang2022posterior}, which adopt $p_{\theta}(\bm{x}|\bm{z})$, $q_{\phi}(\bm{z}|\bm{x})$ and $p_{\rm lat}(\bm z)$ as defined in Eqs.~\eqref{eq:vae_gaussian_pxz}--\eqref{eq:vae_gaussian_pz}, with both $\bm{f}(\bm z)$ and $\bm{\mu}(\bm x)$ taken as linear functions.
Here, we aim to provide an interpretation of posterior collapse as a phase transition, with no limitations on the forms of $\bm{f}(\bm z)$, $\bm{\mu}(\bm x)$ and $\bm{\sigma}(\bm x)$.

As what have discussed before, posterior collapse $q_{\phi}(\bm{z}|\bm{x}) = p_{\rm lat}(\bm z)$ is a trivial solution for VAEs.
Moreover, if we assume the form of $q_{\phi}(\bm{z}|\bm{x})$, such as the choice in Eq.~\eqref{eq:vae_gaussian_pzx}, it is possible that the equation
\begin{equation}
    p_{\rm lat}(\bm z) = \int q_{\phi}(\bm{z}|\bm{x}) p_{\rm data}(\bf x) \dd \bf{x},
\end{equation}
has no solution other than $q_{\phi}(\bm{z}|\bm{x}) = p_{\rm lat}(\bm z)$ under some $p_{\rm data}(\bf x)$~\cite{cremer2018inference}.
Thus, it is reasonable to regard the onset of posterior collapse as a phase transition: from nontrivial, possible multiple extrema points to the trivial one.
Another evidence for claiming posterior collapse as phase transition falls on the KL-divergence term in ELBO.
The KL-divergence $D_{\rm KL}\left(q_{\phi}(\bm{z}|\bm{x})||p_{\rm lat}(\bm z)\right)$ becomes and keeps 0 if fallen into posterior collapse, which may lead to discontinuities in the derivatives of the KL divergence. 

We propose a simple criterion to approximately determine whether posterior collapse occurs for VAEs with the settings as Eqs.~\eqref{eq:vae_gaussian_pxz}--\eqref{eq:vae_gaussian_pz}:
\begin{equation}
    \sigma'^2 > \max\left[\xi_1^2,\cdots,\xi_N^2\right],\label{eq:stability_condition}
\end{equation}
where $\xi_i^2$ ($i=1, 2, \cdots, N$) are the eigenvalues of the covariance matrix $\bm\Sigma$ for the data.
Intuitively, when the decoder noise dominates all principal directions of the data, the latent variables cannot encode meaningful variations, causing the trivial solution to become stable.
We proof this criterion in the Appendix.

We validate the criterion~\eqref{eq:stability_condition} using both synthetic and real-world datasets.
For the synthetic dataset, we use 256,000 samples of 8-dimensional data ($N=8$) drawn from a zero-mean Gaussian distribution, with with eigenvalues of the covariance matrix $\bm{\Sigma}$ given by $\xi_i^2 = \exp\{-(i-1)\lambda\}$ for $i=1,2,\cdots,N$.
We apply the VAE model with two hidden layers in both the encoder (for learning $\bm{\mu}(\bm{x})$ and $\bm{\sigma}(\bm{x})$ in $q_{\phi}(\bm{z}|\bm{x})$) and the decoder (for learning $\bm{f}(\bm{z})$ for $p_{\theta}(\bm{x}|\bm{z})$).
The encoder has hidden layers of dimensions 256 and 128 with ReLU activations, while the decoder has hidden layers of dimensions 128 and 256, also with ReLU activations.
We train and record the trained ELBO, KL-divergence, log likelihood, posterior collapse (Eq.~\eqref{eq:vae_gaussian_extrema}), and mean latent variances $\left<\sigma_j^2\right>$ ($j=1,2,\cdots, n$).

\subsection*{Validation on synthetic dataset}
We set $\lambda = 0.1$, $n=4$, and vary $\sigma'^2$.
The critical point is $\sigma'^2_C =1$.
The trained ELBO converges to the posterior collapse line (Figure. \ref{fig:elbo_n_4}), and $\left<\sigma_j^2\right> = 1$ when $\sigma'^2\geq \sigma'^2_C$ (Figure \ref{fig:var_n_4}).
Furthermore, the KL divergence approaches 0 when $\sigma'^2\geq \sigma'^2_C$ (Figure \ref{fig:kl_n_4}).
We also observe a discontinuity in the derivatives of the KL divergence at $\sigma'^2 = \sigma'^2_C$ by fitting the KL divergence with a piecewise function
\begin{equation}
    f(\sigma'^2)=
    \begin{cases}
        4.22\sigma'^4 - 8.54\sigma'^2 + 4.32 & \left(\sigma'^2 < \sigma'^2_C = 1\right)\\
        0 & \left(\sigma'^2 > \sigma'^2_C = 1\right)
    \end{cases},
\end{equation}
of which the first-order derivative
\begin{equation}
    f'(\sigma'^2) =
    \begin{cases}
        8.44\sigma'^2-8.54 & \left(\sigma'^2 < \sigma'^2_C = 1\right)\\
        0 & \left(\sigma'^2 > \sigma'^2_C = 1\right)
    \end{cases},
\end{equation}
and second-order derivative
\begin{equation}
    f''(\sigma'^2) =
    \begin{cases}
        8.44 & \left(\sigma'^2 < \sigma'^2_C = 1\right)\\
        0 & \left(\sigma'^2 > \sigma'^2_C = 1\right)
    \end{cases},
\end{equation}
show discontinuous.
This discontinuity is indicative of a phase transition in the VAE.

It is important to note that the KL divergence is not interpreted as a thermodynamic functions, such as free energy, in our framework.
Rather, it naturally appears in the ELBO and acts as an order-parameter-like quantity in the non-collapsed phase, where it is strictly positive and varies smoothly with $\sigma'^2$.
The discontinuities in its derivatives at the critical point $\sigma'^2_C$ signal the transition to the collapsed phase, where the KL divergence becomes identically zero.
Thus, it serves as a useful indicator for identifying and characterizing the phase transition associated with posterior collapse.

\begin{figure*}
    \captionsetup{justification=raggedright,singlelinecheck=true}
    \centering
    \begin{subfigure}[b]{0.48\textwidth}
        \centering
        \includegraphics[width=\linewidth]{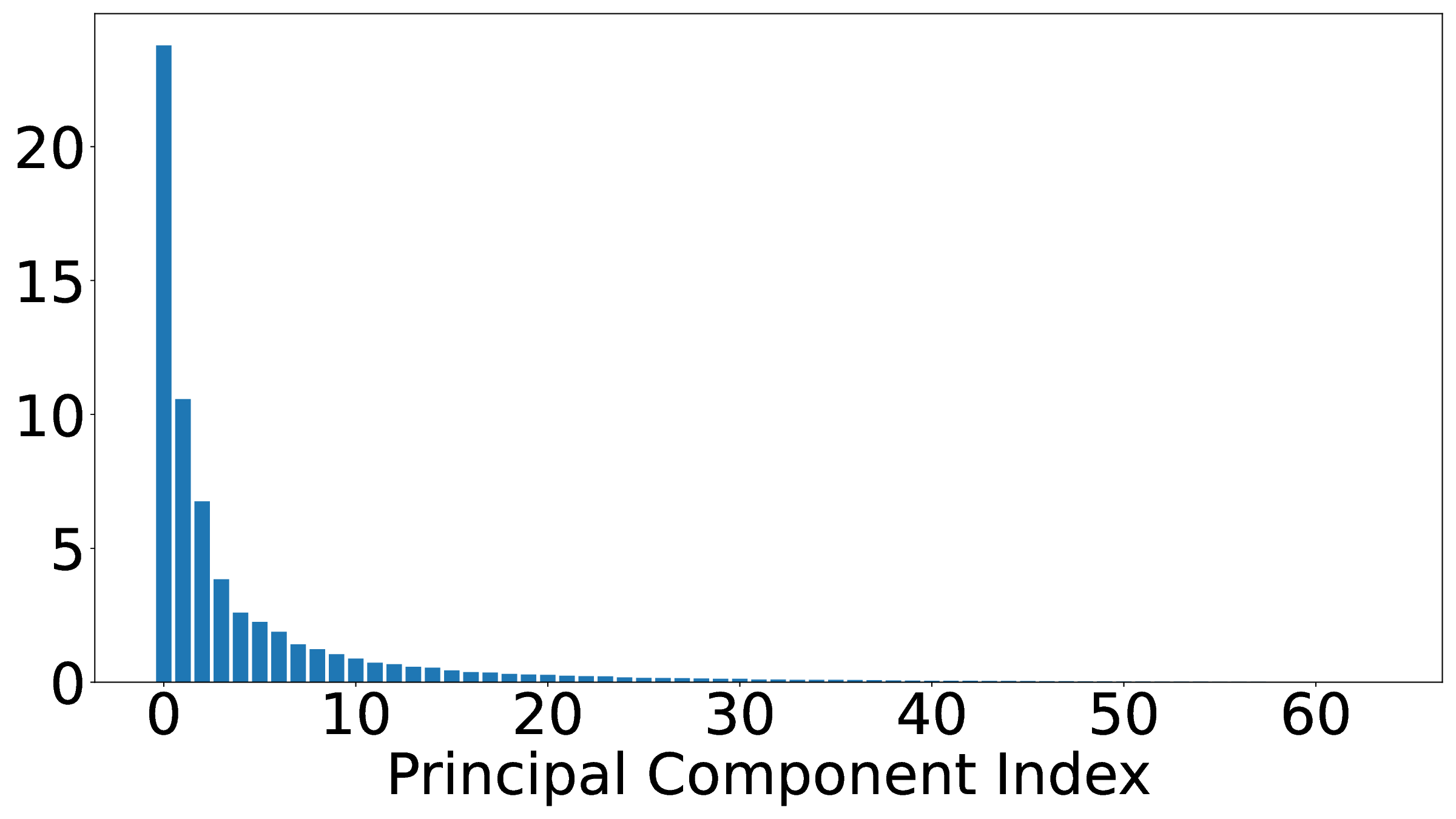}
        \caption{Preprocessed CIFAR10}
        \label{fig:cifar_pca}
    \end{subfigure}
    \hfill
    \begin{subfigure}[b]{0.48\textwidth}
        \centering
        \includegraphics[width=\linewidth]{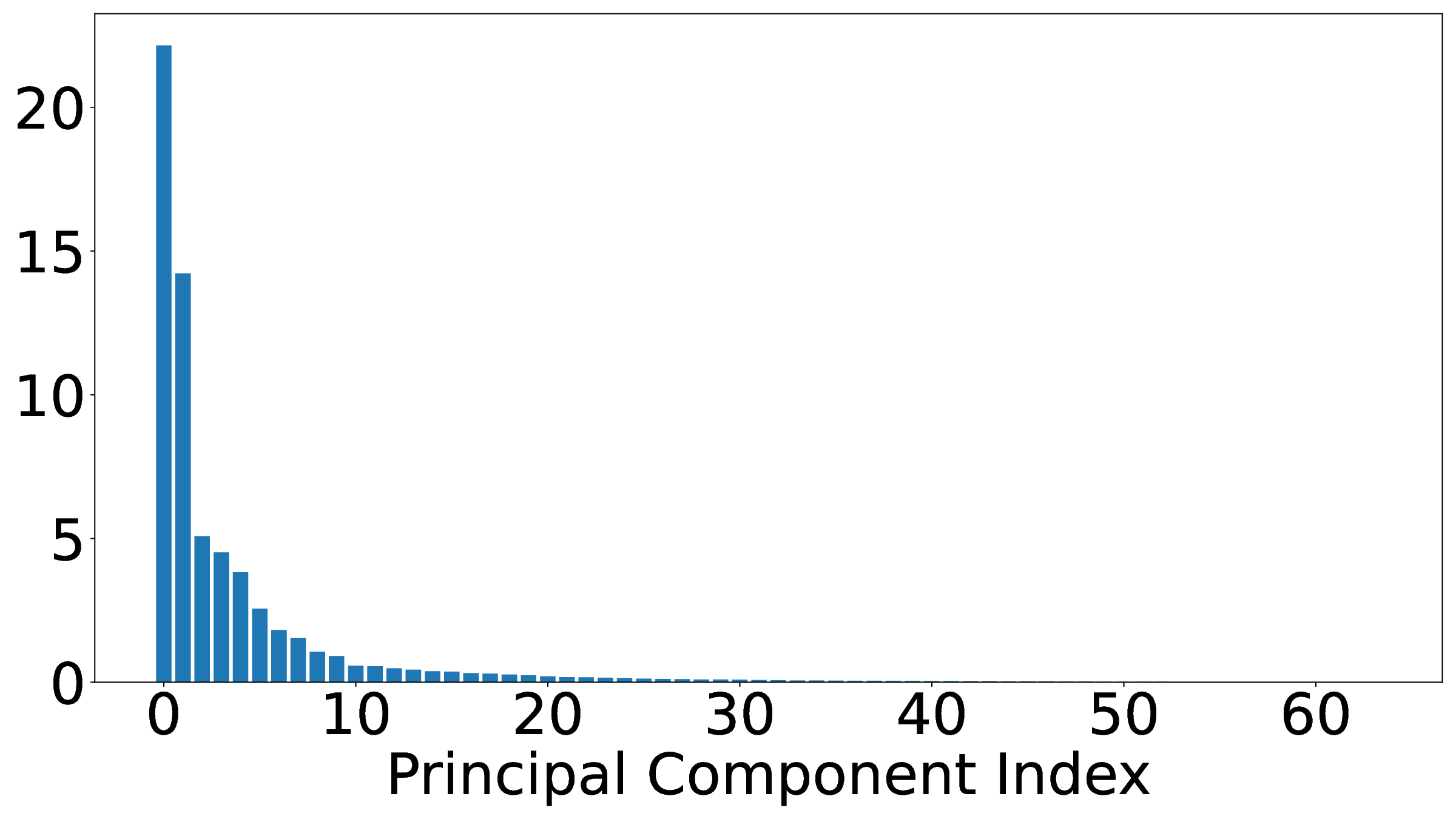}
        \caption{Preprocessed Fashion--MNIST}
        \label{fig:fashion_pca}
    \end{subfigure}
    \caption{(a) Principle components of the preprocessed CIFAR10 data. The maximum one is 23.78, which corresponds to $\sigma'^2_C$.
    (b) Principle components of the preprocessed Fashion-MNIST data. The maximum one is 22.15.}
\end{figure*}

\subsection*{Validation on real-world dataset}
\begin{figure*}
    \captionsetup{justification=raggedright,singlelinecheck=true}
    \centering
    \begin{subfigure}[b]{0.32\textwidth}
        \centering
        \includegraphics[width=\linewidth]{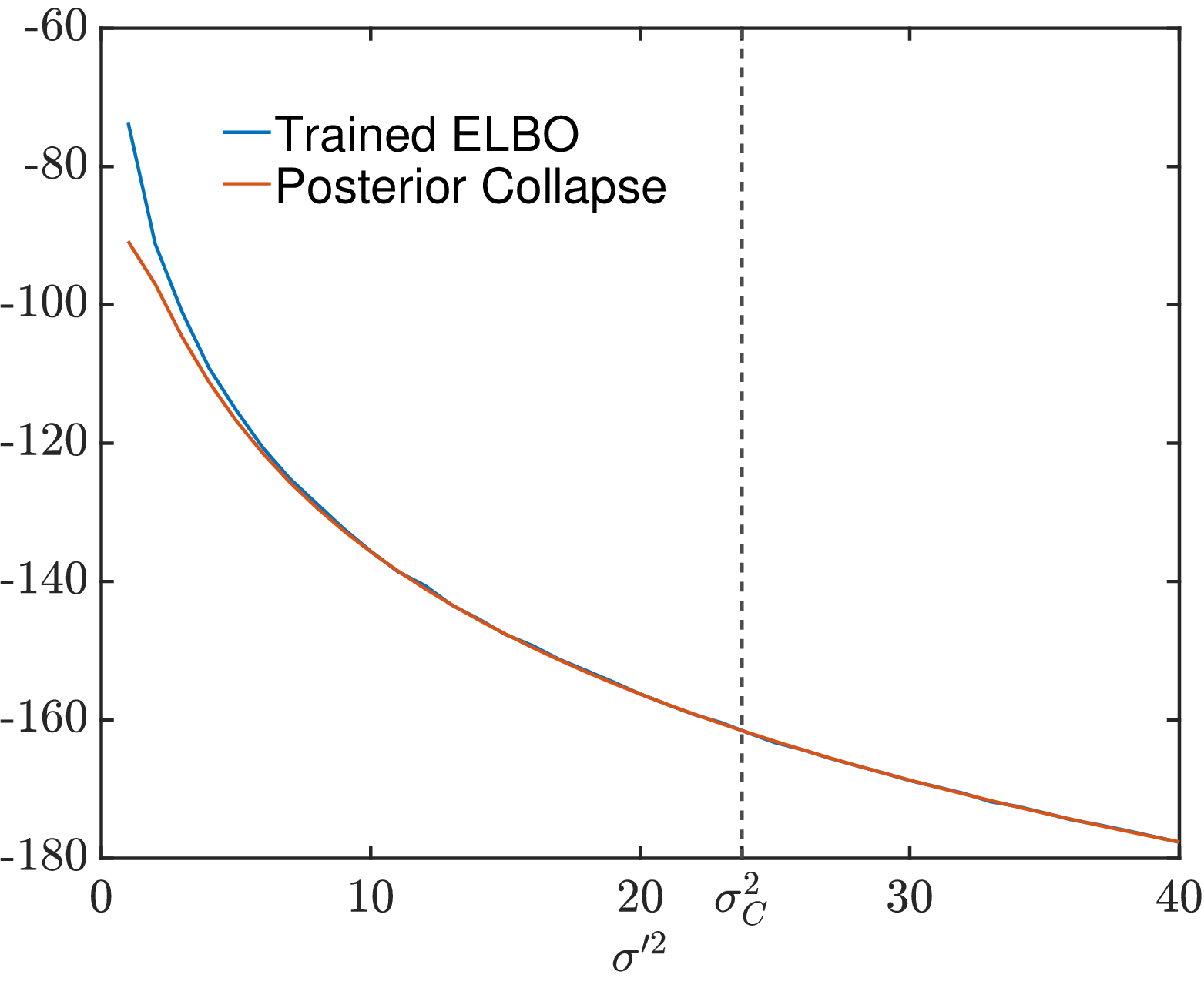}
        \caption{}
        \label{fig:elbo_cifar}
    \end{subfigure}
    \hfill
    \begin{subfigure}[b]{0.32\textwidth}
        \centering
        \includegraphics[width=\linewidth]{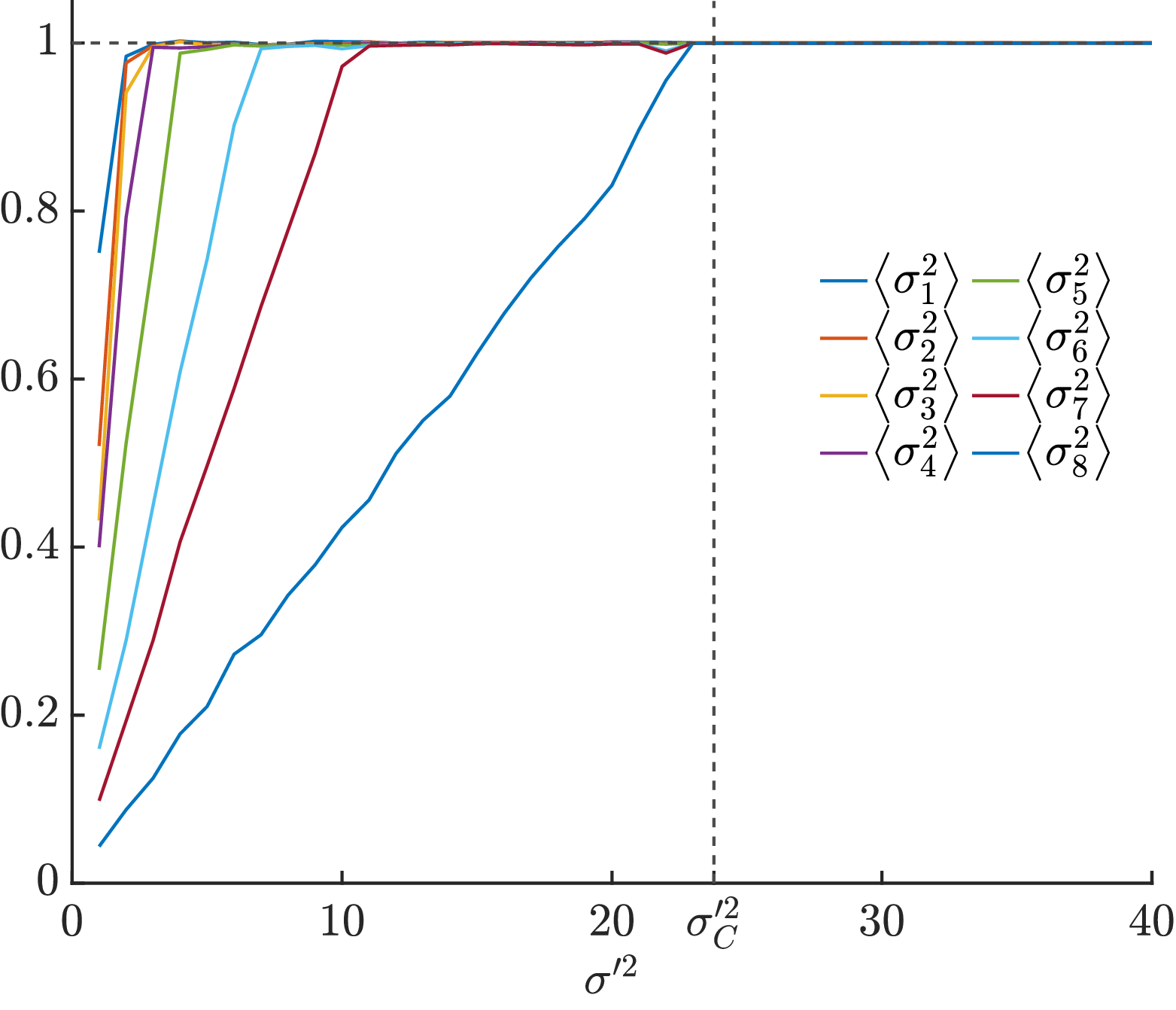}
        \caption{}
        \label{fig:var_cifar}
    \end{subfigure}
    \hfill
    \begin{subfigure}[b]{0.32\textwidth}
        \centering
        \includegraphics[width=\linewidth]{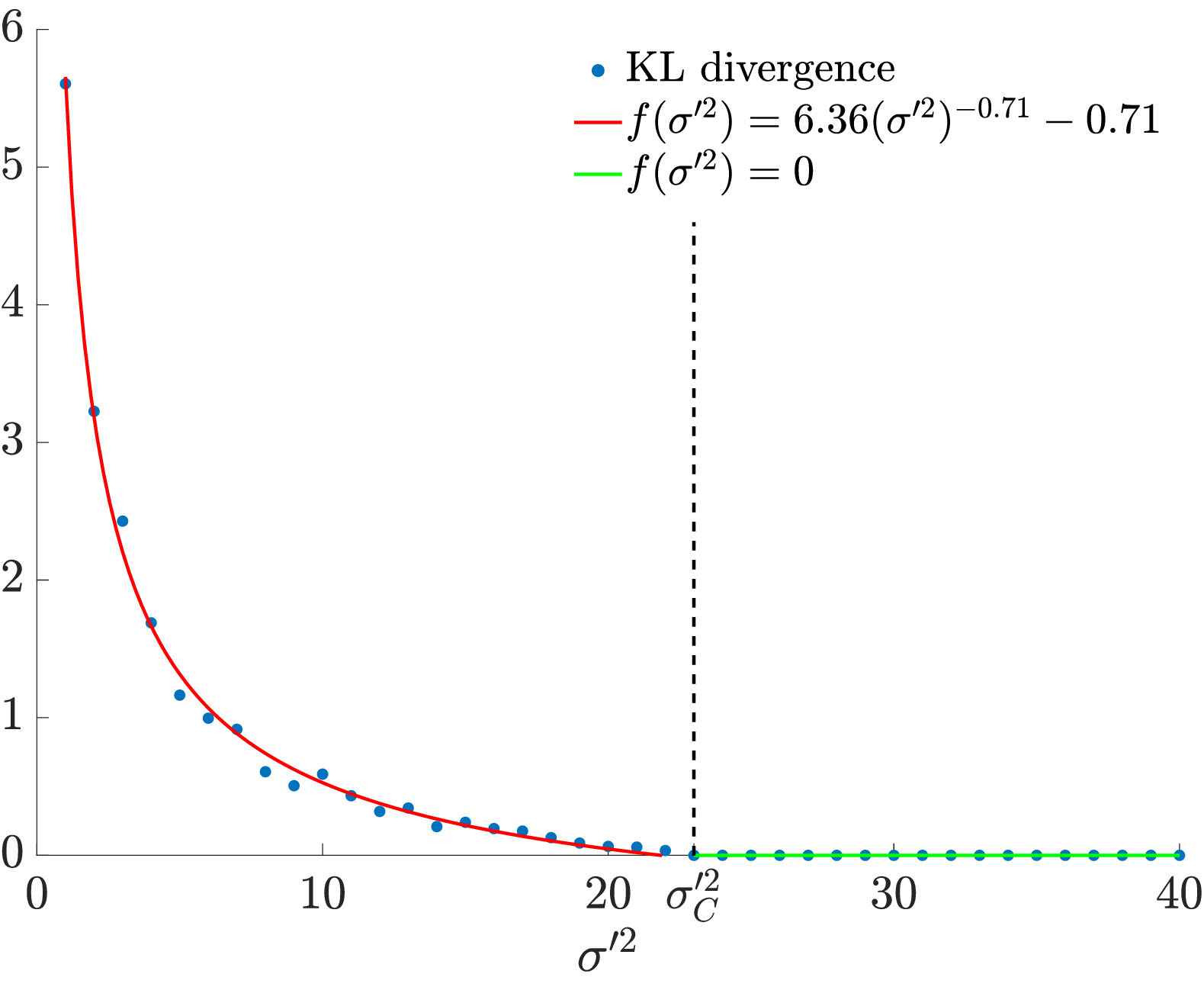}
        \caption{}
        \label{fig:kl_cifar}
    \end{subfigure}
    \caption{(a) Trained ELBO and posterior collapse line v.s. $\sigma'^2$. The trained ELBO converges the posterior collapse when $\sigma'^2>\sigma'^2_C \approx 23.78$.
    (b) Mean latent variances $\left<\sigma_j^2\right>$ ($j=1,\cdots, 8$) v.s. $\sigma'^2$. All of them equal 1 when $\sigma'^2 >\sigma'^2_C\approx 23.78$.
    (c) KL divergence v.s $\sigma'^2$, which converges to 0 when $\sigma'^2>\sigma'^2_C \approx 23.78$.
    The fitted function is used to illustrate the non-analytic derivative behavior near the critical point.}
\end{figure*}

We choose CIFAR10~\cite{Krizhevsky09learningmultiple} as real world data.
CIFAR-10 is a widely used image dataset consisting of 60,000 color images with a resolution of $32\times32$.
To speed up training, we preprocess the dataset by converting color images to grayscale, resizing them to $8\times8$ pixels ($N=64$), and applying standard normalization:
\begin{equation}
    x_i \rightarrow x'_i =\frac{x_i - \left<x_i\right>}{\left<x_i^2\right>-\left<x_i\right>^2}.
\end{equation}
The principle components of $\bm{x}'$ are shown in Figure.~\ref{fig:cifar_pca}, with the largest being 23.78, implying $\sigma'^2_C \approx 23.78$.
The architecture of the VAE model is similar to that used for the synthetic dataset, but the dimensions of hidden layers are 512 and 256 for the encoder, while 256 and 512 for the decoder.
We set $n = 8$ for latent variables.
Similar behaviors are observed near the critical point $\sigma'^2_C \approx 23.78$, as seen in the ELBO (Figure.~\ref{fig:elbo_cifar}), the mean latent variances $\left<\sigma^2_j\right>$ (Figure.~\ref{fig:var_cifar}) and the KL divergence (Figure. \ref{fig:kl_cifar}).
We can also explore the discontinuous of KL-divergence by fitting it with the following piecewise function (red line and green line in Figure.~\ref{fig:kl_cifar}):
\begin{equation}
    f(\sigma'^2)=
    \begin{cases}
        6.36\left(\sigma'^2\right)^{-0.71} - 0.71 & \left(\sigma'^2 < \sigma'^2_C \approx 23.78\right)\\
        0 & \left(\sigma'^2 > \sigma'^2_C \approx 23.78\right)
    \end{cases},
\end{equation}

\subsection*{Validation of $\beta$-VAE and architecture independence}

$\beta$-VAE~\cite{higgins2017beta} introduces a hyper-parameter $\beta$ to balance the reconstruction loss and KL-divergence in the ELBO:
\begin{align}
    \mathcal{L}_{\beta\text{-VAE}} &\equiv \mathbb{E}_{p_{\rm data}(\bm{x})}\left[\mathbb{E}_{\bm{z}\sim q_{\phi}(\bm{z}|\bm{x})}\left[\ln p_{\theta}(\bm{x}|\bm{z})\right]\right]-\beta \mathbb{E}_{p_{\rm data}(\bm{x})}\left[D_{\rm KL}\left(q_{\phi}(\bm{z}|\bm{x}) || p_{\rm lat}(\bm{z})\right)\right].\label{eq:beta_vae_elbo}
\end{align}
By substituting Eqs.~\eqref{eq:vae_gaussian_pxz}--\eqref{eq:vae_gaussian_pz} into Eq.~\eqref{eq:beta_vae_elbo}, we have the training target for deep Gaussian $\beta$-VAEs:
\begin{align}
    \mathcal{L}_{\beta\text{-VAE}}^{\rm G} &\equiv N\ln\frac{1}{\sqrt{2\pi}\sigma'}-\int \frac{1}{2\sigma'^2}\sum_{i=1}^{N}\left[x_i-f_i(\bm{z})\right]^2q_{\phi}(\bm{z}|\bm{x})p_{\rm data}(\bm{x}){\rm d}\bm{x}{\rm d}\bm{z}\notag\\ 
    &\qquad- \frac{\beta}{2}\int\sum_{j=1}^{n}\left[\sigma^2_j(\bm{x})+\mu_j^2(\bm{x})-1-\ln \sigma_j^2(\bm{x})\right]p_{\rm data}(\bm{x}){\rm d}\bm{x},\label{eq:beta_vae_gaussian_elbo}
\end{align}
which takes a similar structure as Eq.~\eqref{eq:vae_gaussian_elbo}.
Indeed, with the replacement
\begin{equation}
    \sigma'^2 \rightarrow \beta \sigma'^2,
\end{equation}
$\mathcal{L}_{\beta\text{-VAE}}^{\rm G}$ can be regarded as $\mathcal{L}_{VAE}^{\rm G}$ for vanilla deep Gaussian VAEs scaling by a factor $\beta$.
This relationship is also implied in previous analysis of linear VAEs~\cite{dang2023beyond,wang2022posterior}.
Thus, the criterion for posterior collapse in $\beta$-VAE reads:
\begin{equation}
    \beta \sigma'^2 > \max\left[\xi_1^2,\cdots,\xi_N^2\right].\label{eq:beta_vae_stability_condition}
\end{equation}

We validate the criterion~\eqref{eq:beta_vae_stability_condition} using Fashion-MNIST~\cite{xiao2017fashion}, which consists of 70,000 grayscale images of size $28\times28$ ($N=784$).
We preprocess the dataset with the same method in CIFAR10.
The principle components are shown in Figure.~\ref{fig:fashion_pca}, with the largest being 22.15.

\begin{figure*}
    \captionsetup{justification=raggedright,singlelinecheck=true}
    \centering
    \begin{subfigure}[b]{0.32\textwidth}
        \centering
        \includegraphics[width=\linewidth]{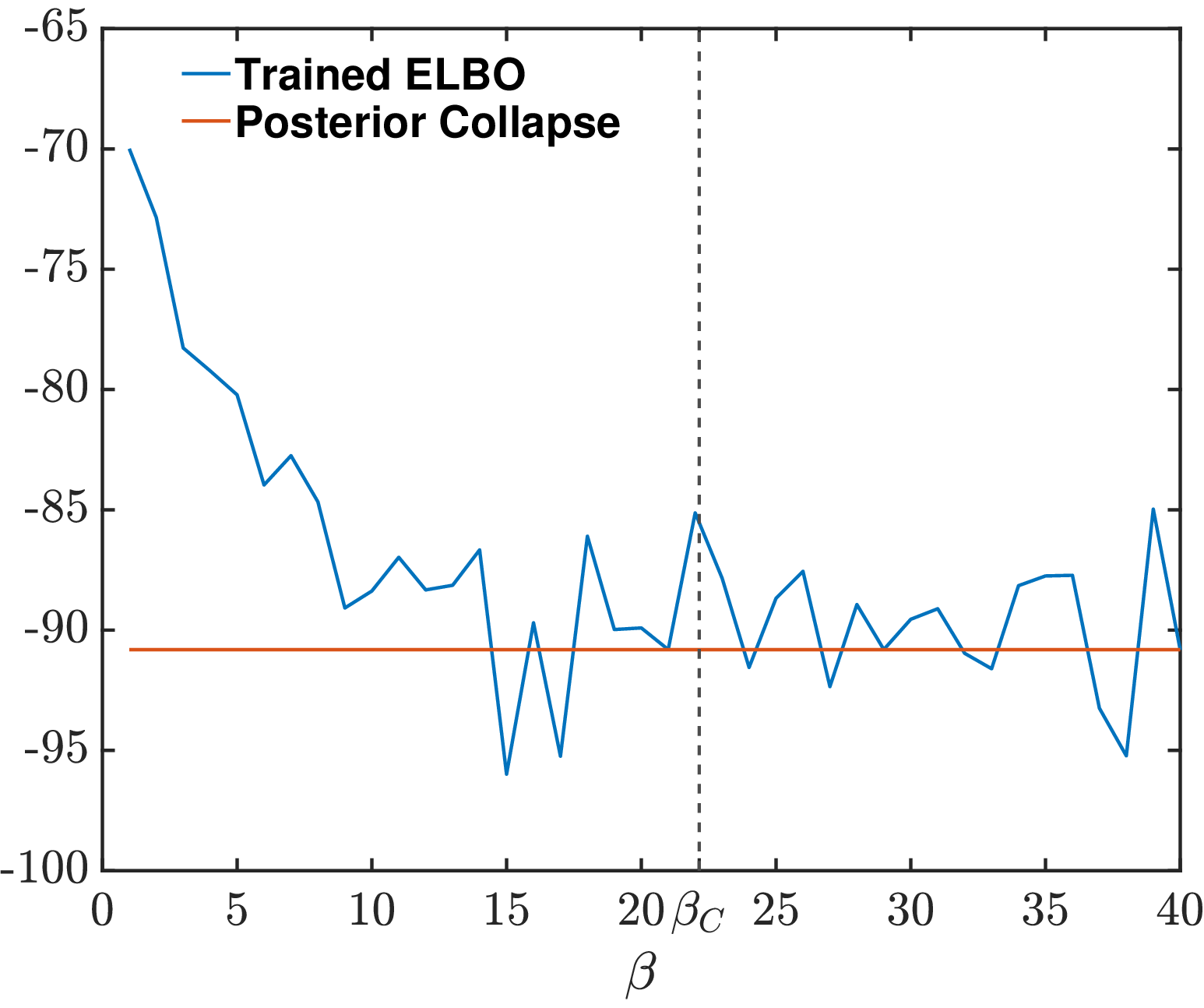}
        \caption{}
        \label{fig:elbo_fashion}
    \end{subfigure}
    \hfill
    \begin{subfigure}[b]{0.32\textwidth}
        \centering
        \includegraphics[width=\linewidth]{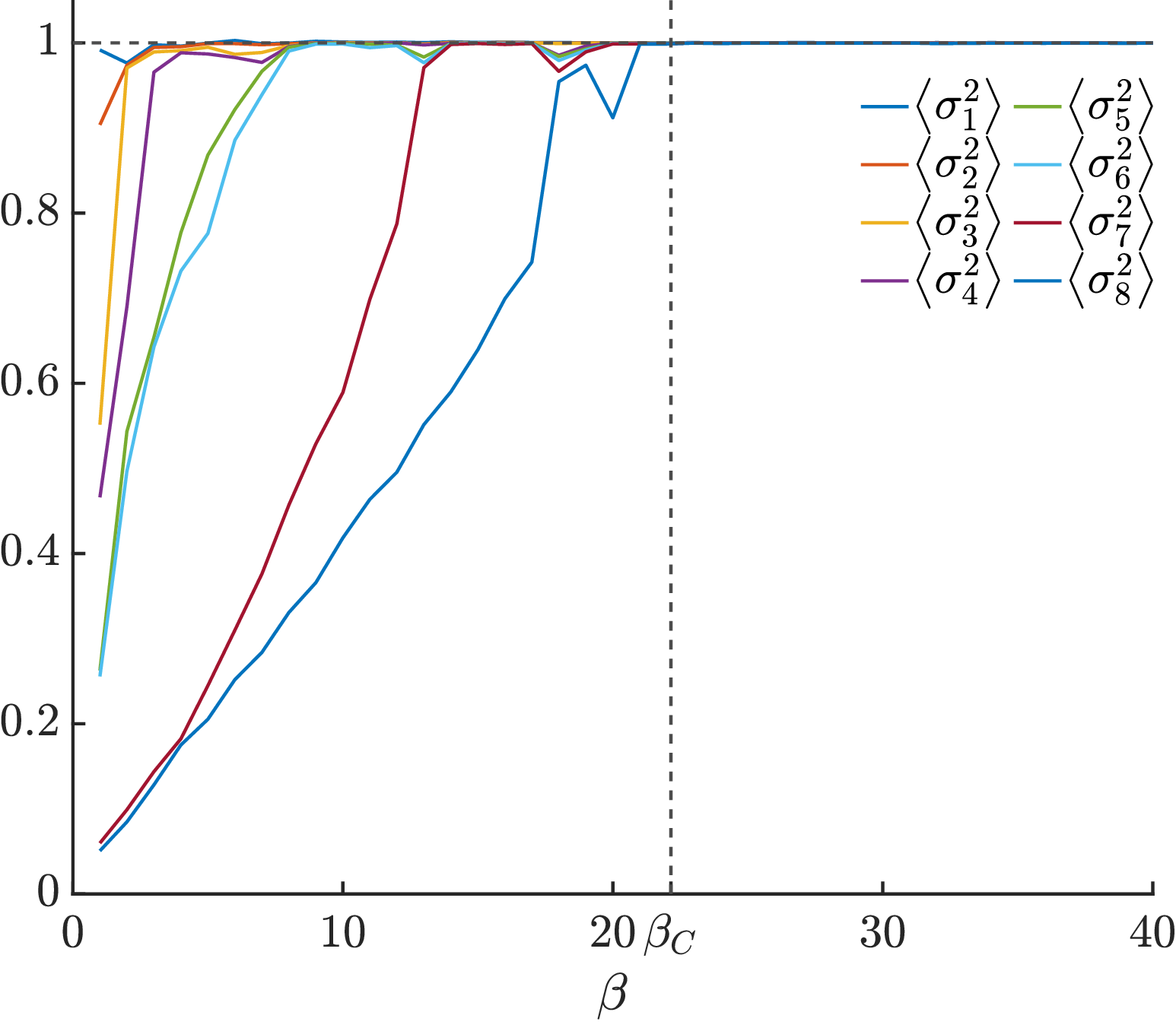}
        \caption{}
        \label{fig:var_fashion}
    \end{subfigure}
    \hfill
    \begin{subfigure}[b]{0.32\textwidth}
        \centering
        \includegraphics[width=\linewidth]{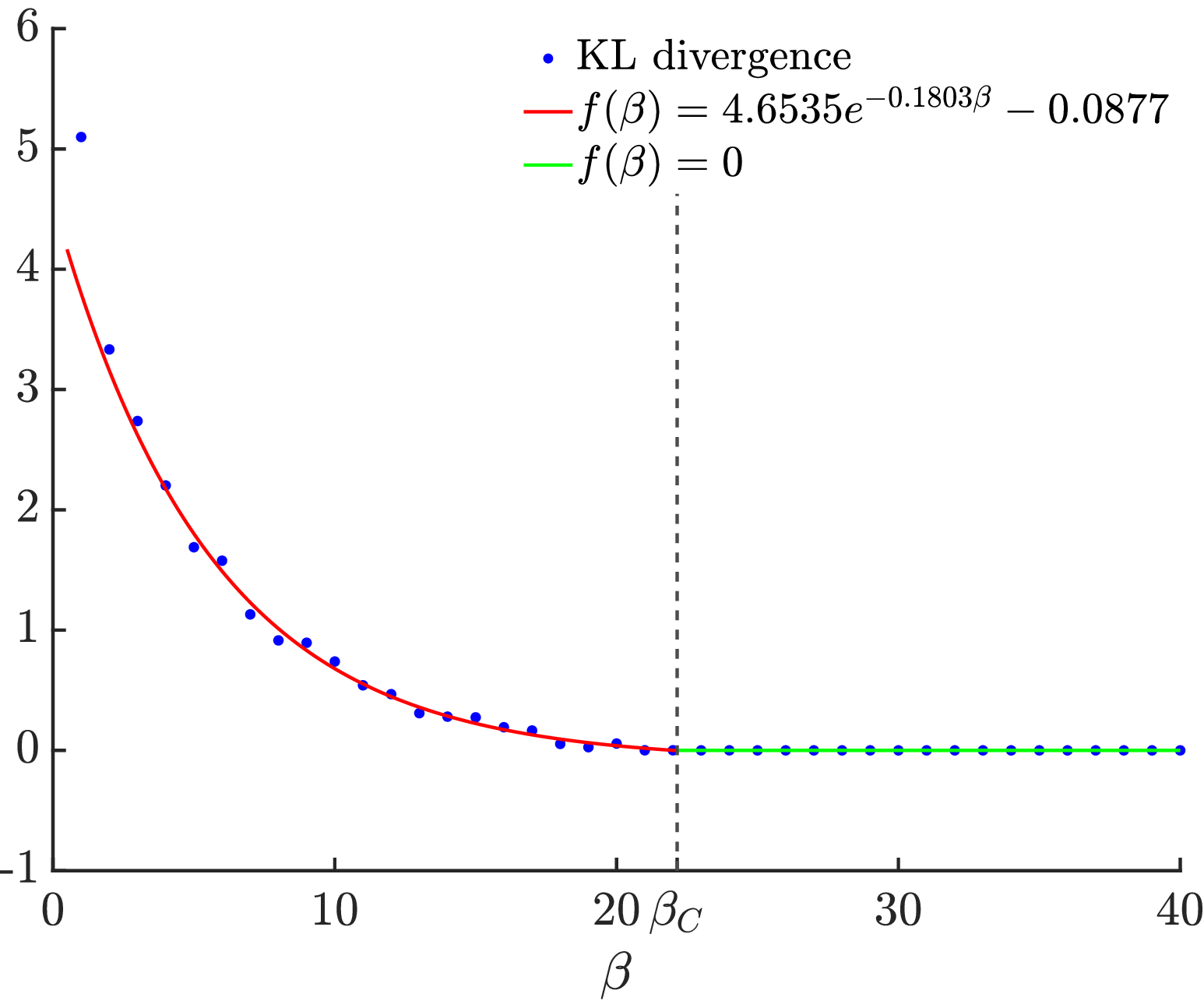}
        \caption{}
        \label{fig:kl_fashion}
    \end{subfigure}
    \caption{(a) Trained ELBO and posterior collapse line v.s. $\beta$. The trained ELBO converges the posterior collapse when $\beta>\beta_C \approx 22.15$.
    (b) Mean latent variances $\left<\sigma_j^2\right>$ ($j=1,\cdots, 8$) v.s. $\beta$. All of them equal 1 when $\beta >\beta_C\approx 22.15$.
    (c) KL divergence v.s $\beta$, which converges to 0 when $\beta>\beta_C \approx 22.15$.
    The fitted function is used to illustrate the non-analytic derivative behavior near the critical point.}
\end{figure*}

To demonstrate that our criterion, Eq.~\eqref{eq:stability_condition} or~\eqref{eq:beta_vae_stability_condition}, does not depend on the specific neural
architecture, we employ convolutional encoder and decoder networks in this $\beta$-VAE.
The encoder includes two convolutional layers with ReLU activations, while the decoder consists of two transposed convolutional layers, also activated with ReLU.
We set $n=8$ for latent variables and $\sigma'^2=1$, making the critical point $\beta_C \approx 22.15$.
Similar behaviors are observed near the critical point $\beta_C \approx 22.15$, as seen in the ELBO (Figure.~\ref{fig:elbo_fashion}), the mean latent variances $\left<\sigma^2_j\right>$ (Figure.~\ref{fig:var_fashion}) and the KL divergence (Figure. \ref{fig:kl_fashion}).
We can also explore the discontinuous of KL-divergence by fitting it with the following piecewise function (red line and green line in Figure.~\ref{fig:kl_fashion}):
\begin{equation}
    f(\beta)=
    \begin{cases}
        4.65e^{-0.18\beta} - 0.09 & \left(\beta < \beta_C \approx 22.15\right)\\
        0 & \left(\beta > \beta_C \approx 22.15\right)
    \end{cases},
\end{equation}
of which the derivatives also show discontinuities at $\beta = \beta_C$.

\section{Applicability to real-world problems}
\label{sec_applicability}

Although the analysis developed in this work is rooted in the mathematical structure 
of the ELBO, the resulting collapse criterion has several direct implications for 
practical machine learning tasks involving VAEs.

\subsection*{Hyperparameter selection and model design}

The collapse boundary~\eqref{eq:stability_condition} provides a principled guideline for avoiding posterior collapse.
In practice, this means that a practitioner may inspect the covariance spectrum of a dataset and select decoder noise levels or KL regularization strengths that ensure the stability of the informative solution. 
This offers a diagnostic tool for determining when collapse is likely to appear 
during training, without the need for extensive trial-and-error hyperparameter search.

\subsection*{Applicability to non-linear and convolutional architectures}

Because our stability analysis depends only on the local behavior of the ELBO  around the trivial extremum, the resulting criterion is independent of the specific parameterization of the encoder and decoder networks. 
This includes widely used non-linear architectures such as convolutional VAEs in image modeling and anomaly detection. 
Our empirical results confirm that the predicted collapse threshold closely matches the observed behavior even when the encoder and decoder are highly expressive neural networks.

\subsection*{Relevance to practical domains}

Posterior collapse is a common bottleneck in real-world generative modeling tasks, including representation learning, semi-supervised learning, image synthesis, and industrial anomaly detection. 
The stability-based framework presented here offers a quantitative interpretation of when these failures occur, and provides insight into how the architecture or hyperparameters may be adjusted to maintain informative latent representations.

\section{Conclusions}
\label{sec_con}
In this work, we have established a new perspective on posterior collapse in VAEs by interpreting it as a phase transition driven by the interplay between data structure and model hyper-parameters.
By analyzing the stability of the trivial solution corresponding to posterior collapse, we derived a critical point characterized by the spectral properties of the data covariance.
This condition predicts the onset of collapse, which we validated on both constructed and real world datasets.

Our findings suggest that posterior collapse is not merely an artifact of optimization dynamics but a fundamental emergent behavior governed by statistical and variational constraints.
This perspective not only bridges variational learning with phase transition theory but also offers practical tools for preventing posterior collapse in VAE training.
The KL divergence therefore provides a convenient empirical indicator of the transition: although it becomes exactly zero in the collapsed phase, the one-sided derivatives approaching the critical point from the non-trivial phase encode the non-analytic structure characteristic of phase transitions.

Importantly, the theoretical criterion we derive is consistent with prior results in linear VAEs~\cite{lucas2019don, dang2023beyond, wang2022posterior}, as the stability of the collapse point is mainly determined by linear perturbations of the encoder and decoder functions.
Moreover, beyond the bad local optima~\cite{dai2020usual}, our analysis offers a complementary perspective by identifying posterior collapse as a kind of phase transition that reflects global data-model interactions.

As we have mentioned in Sec.~\ref{sec_post}, the role of the reconstruction noise variance $\sigma'^2$ is analogous to the $\beta$ parameter in $\beta$-VAE, adjusting the weights of the reconstruction loss and KL-divergence in the ELBO.
Thus, introducing a $\beta$ coefficient for the KL-divergence is equivalent to rescaling the ELBO and $\sigma'^2$, and shifting the critical point accordingly.

Future work could extend the current analysis to settings with non-Gaussian data distributions or decoders, where the reconstruction loss no longer has a simple quadratic form.
Such extensions could reveal new types of phase behavior and potentially uncover richer collapse dynamics influenced by the shape and tails of the data distribution.
In addition, it is worth investigating the generality of this phase transition framework in extended architectures such as conditional or hierarchical VAEs, and further explore its implications in understanding the expressivity and limitations of deep generative models.

Moreover, it is also worth exploring more precious theories to identify the posterior collapse as well as analysis the order of the phase transition, beyond discussing the stability of the trivial solutions~\eqref{eq:extrema_1}--\eqref{eq:extrema_3}.
Such theories may require analytical solutions of extremes other than the trivial one, and the corresponding stability.

Another promising direction is to investigate how posterior collapse emerges dynamically during training.
As suggested by Figures~\ref{fig:var_n_4} and~\ref{fig:var_cifar}, and previous works on linear models~\cite{lucas2019don, dang2023beyond, wang2022posterior}, the collapse of the variational posterior may not occur simultaneously across all latent dimensions, but rather in a sequential manner.
Specifically, each latent variance $\left<\sigma^2_j\right>$ appears to approach 1 at different choice of the hyper-parameter $\sigma'^2$, implying that posterior collapse may form via the successive destabilization of individual latent directions.
This observation raises the possibility of a more nuanced collapse mechanism governed by the alignment of latent directions with the principal components of the data.
Developing a refined theoretical framework that accounts for this dimension-wise collapse could provide deeper insight into the onset of collapse and suggest new strategies—such as targeted regularization or selective annealing—to mitigate it and preserve informative latent structure.

\section{Acknowledgements}
This work was supported by JST SPRING, Grant Number JPMJSP2108.

\appendix
\section{Derivation criterion~\eqref{eq:stability_condition}}
Criterion~\eqref{eq:stability_condition} comes from analyzing the stability of the trivial extremum given by~\eqref{eq:extrema_1}--\eqref{eq:extrema_3}.
If the trivial local extrema of $\mathcal{L}^G_{\rm VAE}$ is unstable, the VAE may escape posterior collapse; otherwise, it may converge to the collapsed solution.
We analyze this stability with the following framework.

The second order variation of $\mathcal{L}_{\rm VAE}^G$ in Eq.~\eqref{eq:vae_gaussian_elbo} at the trivial extrema point~\eqref{eq:extrema_1}--\eqref{eq:extrema_3} reads:
\begin{equation}
    \delta^2 \mathcal{L}_{\rm VAE}^G = \int (\delta \bm{f}, \delta\bm\mu, \delta\bm\sigma) \bm H^G (\delta \bm{f}, \delta\bm\mu, \delta\bm\sigma)^\top \dd \tilde{\bm x} \dd \bm z,\label{eq:2nd_variation}
\end{equation}
where
\begin{align}
    \bm H^G &= 
    \begin{pmatrix}
        \bm A & \bm D & \bm E\\
        \bm D^\top & \bm B & \bm 0\\
        \bm E^\top & \bm 0 & \bm C
    \end{pmatrix},\\
    A_{ij} & = -\frac{\delta_{ij}}{\sigma'^2}\tilde{p}_{\rm data}(\tilde{\bm x})p_{\rm lat}(\bm z),\\
    B_{ij} & = -\delta_{ij}\tilde{p}_{\rm data}(\tilde{\bm x})p_{\rm lat}(\bm z),\\
    C_{ij} & = -2\delta_{ij}\tilde{p}_{\rm data}(\tilde{\bm x})p_{\rm lat}(\bm z),\\
    D_{ij} & = \frac{1}{\sigma'^2}\left[\sum_{k=1}^{N}\left(\bm{U}^{-1}\right)_{ik}\xi_k\tilde{x}_k\right] z_j \tilde{p}_{\rm data}(\tilde{\bm x})p_{\rm lat}(\bm z),\\
    E_{ij} & = \frac{1}{\sigma'^2}\left[\sum_{k=1}^{N}\left(\bm{U}^{-1}\right)_{ik}\xi_k\tilde{x}_k\right](z_j^2 - 1) \tilde{p}_{\rm data}(\tilde{\bm x})p_{\rm lat}(\bm z).
\end{align}
Here, we perform the transformation (PCA whitening~\cite{kessy2018optimal})
\begin{equation}
    x_i\rightarrow\tilde{x}_i = \sum_{j=1}^N U_{ij}\left({x_j - \left<x_j\right>}\right)/{\xi_i},
\end{equation}
and correspondingly,
\begin{equation}
    p_{\rm data}(\bm x) \rightarrow \tilde{p}_{\rm data}(\tilde{\bm x}) = \prod_{i=1}^N \frac{1}{\sqrt{2\pi}}\exp\left(-\frac{\tilde{x}_i^2}{2}\right).
\end{equation}
The orthogonal matrix $\bm U$ ($\bm U^{-1} = \bm U^\top$) transforms the data $\bm x$ into the directions of its principal components (eigenvalues of the covariance matrix) $\xi_i^2$ ($i=1,2,\cdots,N$).  
The trivial extrema point is stable if $\delta^2 \mathcal{L}_{\rm VAE}^G < 0$.

We expand $\delta \bm{f}(\bm z)$, $\delta\bm\mu(\tilde{\bm x})$ and $\delta\bm\sigma(\tilde{\bm x})$ in terms of Hermite polynomials $H_n$:
\begin{align}
    \delta f_i(\bm z) &= \sum_{m_1,\cdots,m_n}\alpha^i_{m_1,\cdots,m_n}H_{m_1}(z_1)\cdots H_{m_1}(z_n),\\
    \delta \mu_j(\tilde{\bm x}) &= \sum_{m_1,\cdots,m_N}\beta^j_{m_1,\cdots,m_N}H_{m_1}(\tilde{x}_1)\cdots H_{m_N}(\tilde{x}_N),\\
    \delta \sigma_j(\tilde{\bm x}) &= \sum_{m_1,\cdots,m_N}\gamma^j_{m_1,\cdots,m_N}H_{m_1}(\tilde{x}_1)\cdots H_{m_N}(\tilde{x}_N).
\end{align}
We choose Hermite polynomials
\begin{equation}
    H_n (x) = (-1)^n e^{{x^2}/{2}}\frac{d^n}{dx^n}e^{{-x^2}/{2}},
\end{equation}
due to their orthogonality properties~\cite{fai2024special}
\begin{equation}
    \int H_m(x)H_n(x)\frac{1}{\sqrt{2\pi}}e^{-x^2/2} \dd x = n!\delta_{mn}, 
\end{equation}
and the recurrence relations~\cite{fai2024special}
\begin{align}
    xH_n(x) &=H_{n+1}(x)+nH_{n-1}(x),\\
    x^2 H_n(x) &= H_{n+2}(x)+(2n+1)H_n(x)+n(n-1)H_{n-2}(x).
\end{align}

Then, we can write each part of the integral in Eq.~\eqref{eq:2nd_variation} as:
\begin{align}
    \int A_{ij} \delta f_i \delta f_j \dd \tilde{\bm x} \dd \bm z&= -\frac{1}{\sigma'^2}m_1!\cdots m_n!\left(\alpha^i_{m_1,\cdots,m_n}\right)^2,\\
    \int B_{jk} \delta \mu_j \delta \mu_k \dd \tilde{\bm x} \dd \bm z&= -m_1!\cdots m_N!\left(\beta^j_{m_1,\cdots,m_N}\right)^2,\\
    \int C_{jk} \delta \sigma_j \delta \sigma_k \dd \tilde{\bm x} \dd \bm z&= -2m_1!\cdots m_N!\left(\gamma^j_{m_1,\cdots,m_N}\right)^2,\\
    \int D_{ij} \delta f_i \delta \mu_j \dd \tilde{\bm x} \dd \bm z&= \frac{1}{\sigma'^2}\left[\sum_{k=1}^{N}\left(\bm{U}^{-1}\right)_{ik}\xi_k\right] \alpha^i_{m_j=1}\beta^j_{m_k=1},\\
    \int E_{ij} \delta f_i \delta \mu_j \dd \tilde{\bm x} \dd \bm z&= \frac{2}{\sigma'^2}\left[\sum_{k=1}^{N}\left(\bm{U}^{-1}\right)_{ik}\xi_k\right] \alpha^i_{m_j=2}\gamma^j_{m_k=1},
\end{align}
and calculate $\delta^2\mathcal{L}_{\rm VAE}^G$.
Here, we use $\alpha^i_{m_j=1}$ to denote $\alpha^i_{m_1,\cdots,m_n}$ if $m_j = 1$ and all other $m_i = 0$ for $i\neq j$.
The same convention applies to $\alpha^i_{m_j=2}$, $\beta^j_{m_i=1}$ and $\gamma^j_{m_i=1}$.
Thus, ensuring the negative definiteness of $\delta^2\mathcal{L}_{\rm VAE}^G$ reduces to checking the negative definiteness of the following quadratic expressions:
\begin{align}
    &-\frac{1}{\sigma'^2}\sum_{i=1}^{N}\sum_{j=1}^n\left(\alpha^i_{m_j = 1}\right)^2-\sum_{i=1}^{N}\sum_{j=1}^n\left(\beta^j_{m_i = 1}\right)^2 + \frac{2}{\sigma'^2}\sum_{i,k=1}^{N}\sum_{j=1}^n\left(\bm{U}^{-1}\right)_{ik}\xi_k \alpha^i_{m_j = 1}\beta^j_{m_k = 1},\label{eq:bino1}\\
    &-\frac{2}{\sigma'^2}\sum_{i=1}^{N}\sum_{j=1}^n\left(\alpha^i_{m_j = 2}\right)^2-2\sum_{i=1}^{N}\sum_{j=1}^n\left(\gamma^j_{m_i = 1}\right)^2 + \frac{4}{\sigma'^2}\sum_{i,k=1}^{N}\sum_{j=1}^n\left(\bm{U}^{-1}\right)_{ik}\xi_k \alpha^i_{m_j = 2}\gamma^j_{m_k = 1},\label{eq:bino2}
\end{align}
both of which lead to the same condition.

Take the quadratic expression~\eqref{eq:bino1} for example, to derive the condition for negative definiteness, we denote vector $\bm \alpha^*$ and $\bm \beta^*$ as
\begin{align}
    \alpha^*_{i+(j-1)n} &\equiv \alpha^i_{m_j=1},\\
    \beta^*_{i+(j-1)n} &\equiv \beta^j_{m_i=1}.
\end{align}
We also denote $\bm \Lambda = \mathrm{diag}(\xi_1, \cdots, \xi_N)$.
Further, we can write the quadratic expression~\eqref{eq:bino1} as a matrix form
\begin{equation}
    -
    \begin{pmatrix}
        \bm \alpha^{*\top}, \bm \beta^{*\top}
    \end{pmatrix}
    \bm M
    \begin{pmatrix}
        \bm \alpha\\
        \bm \beta
    \end{pmatrix},
\end{equation}
where
\begin{equation}
    \bm M \equiv\sigma'^{-2}\begin{pmatrix}
        \bm I_{Nn} & -\left[\bm I_{n}\otimes\left(\bm U^{-1}\bm \Lambda\right)\right]\\
        -\left[\bm I_{n}\otimes\left(\bm U^{-1}\bm \Lambda\right)\right]^\top & \sigma'^{2}\bm I_{Nn}
    \end{pmatrix},
\end{equation}
$I$ is the identity matrix, and $\otimes$ refers to the 
Kronecker product.
As $\bm I_{Nn}$ is positive definite, the positive definite of its Schur complement~\cite{zhang2006schur} $\bm S$:
\begin{align}
    \bm S &= \sigma'^2\bm I_{Nn} - \left[\bm I_{n}\otimes\left(\bm U^{-1}\bm \Lambda\right)\right]^\top\left[\bm I_{n}\otimes\left(\bm U^{-1}\bm \Lambda\right)\right]\notag\\
    &= \sigma'^2\bm I_{Nn} - \left[\bm I_{n}\otimes\left(\bm\Lambda^\top(\bm U^{-1})^\top \bm U^{-1}\bm \Lambda\right)\right]\notag\\
    &= \sigma'^2\bm I_{Nn} - \left[\bm I_{n}\otimes\left(\bm\Lambda^\top\bm\Lambda\right)\right]\notag\\
    &= \bm I_{n} \otimes \mathrm{diag}\left(\sigma'^2-\xi_1^2,\cdots, \sigma'^2-\xi_N^2\right),
\end{align}
leads to the positive definiteness of $\bm M$ and the negative definiteness of the quadratic expression~\eqref{eq:bino1}, which requires the criterion~\eqref{eq:stability_condition}:
\begin{equation}
    \sigma'^2 > \max\left[\xi_1^2,\cdots,\xi_N^2\right].
\end{equation}

The criterion is consistent with results obtained for linear VAEs~\cite{lucas2019don,dang2023beyond,wang2022posterior}.
Binomials~\eqref{eq:bino1} and \eqref{eq:bino2} requires the same condition to be negative definiteness.
$\alpha^i_{m_j = 1}$ and $\beta^j_{m_i = 1}$ in binominals~\eqref{eq:bino1} describes the linear perturbation ($H_1$ is linear) of $f_i(\tilde{\bm z})$ and $\mu_j(\tilde{\bm x})$ from the trivial extrema point respectively, which corresponds to the linear VAE setting where $\bm f(\bm z) = \bm{Vz}+\bm{b}$ and $\bm \mu(\bm x) = \bm{Wx}+\bm{c}$~\cite{lucas2019don,dang2023beyond,wang2022posterior}.

\bibliographystyle{elsarticle-num.bst}

\begin{thebibliography}{10}
\expandafter\ifx\csname url\endcsname\relax
  \def\url#1{\texttt{#1}}\fi
\expandafter\ifx\csname urlprefix\endcsname\relax\def\urlprefix{URL }\fi
\expandafter\ifx\csname href\endcsname\relax
  \def\href#1#2{#2} \def\path#1{#1}\fi

\bibitem{kingma2013auto}
D.~P. Kingma, M.~Welling, et~al., \href{https://arxiv.org/abs/1312.6114}{Auto-encoding variational bayes} (2013).

\bibitem{girin2020dynamical}
L.~Girin, S.~Leglaive, X.~Bie, J.~Diard, T.~Hueber, X.~Alameda-Pineda, \href{https://arxiv.org/abs/2008.12595}{Dynamical variational autoencoders: A comprehensive review}, arXiv preprint arXiv:2008.12595 (2020).

\bibitem{blei2017variational}
D.~M. Blei, A.~Kucukelbir, J.~D. McAuliffe, \href{https://www.tandfonline.com/doi/full/10.1080/01621459.2017.1285773}{Variational inference: A review for statisticians}, Journal of the American Statistical Association 112~(518) (2017) 859--877.

\bibitem{jordan1999introduction}
M.~I. Jordan, Z.~Ghahramani, T.~S. Jaakkola, L.~K. Saul, \href{https://link.springer.com/article/10.1023/A:1007665907178}{An introduction to variational methods for graphical models}, Machine Learning 37 (1999) 183--233.

\bibitem{huang2018introvae}
H.~Huang, R.~He, Z.~Sun, T.~Tan, et~al., \href{https://proceedings.neurips.cc/paper/2018/hash/093f65e080a295f8076b1c5722a46aa2-Abstract.html}{Introvae: Introspective variational autoencoders for photographic image synthesis}, Advances in Neural Information Processing Systems 31 (2018).

\bibitem{lin2020anomaly}
S.~Lin, R.~Clark, R.~Birke, S.~Sch{\"o}nborn, N.~Trigoni, S.~Roberts, \href{https://ieeexplore.ieee.org/abstract/document/9053558}{Anomaly detection for time series using vae-lstm hybrid model}, in: ICASSP 2020-2020 IEEE International Conference on Acoustics, Speech and Signal Processing (ICASSP), IEEE, 2020, pp. 4322--4326.

\bibitem{dollar2023efficient}
O.~Dollar, N.~Joshi, J.~Pfaendtner, D.~A. Beck, \href{https://pubs.acs.org/doi/full/10.1021/acs.jpca.3c04188}{Efficient 3d molecular design with an e (3) invariant transformer vae}, The Journal of Physical Chemistry A 127~(37) (2023) 7844--7852.

\bibitem{lucas2019don}
J.~Lucas, G.~Tucker, R.~B. Grosse, M.~Norouzi, \href{https://proceedings.neurips.cc/paper/2019/hash/7e3315fe390974fcf25e44a9445bd821-Abstract.html}{Don't blame the elbo! a linear vae perspective on posterior collapse}, Advances in Neural Information Processing Systems 32 (2019).

\bibitem{dang2023beyond}
H.~Dang, T.~Tran, T.~Nguyen, N.~Ho, \href{https://arxiv.org/abs/2306.05023}{Beyond vanilla variational autoencoders: Detecting posterior collapse in conditional and hierarchical variational autoencoders}, arXiv preprint arXiv:2306.05023 (2023).

\bibitem{wang2022posterior}
Z.~Wang, Z.~Liu, \href{https://openreview.net/forum?id=zAc2a6_0aHb}{Posterior collapse of a linear latent variable model}, Advances in Neural Information Processing Systems 35 (2022) 37537--37548.

\bibitem{dai2020usual}
B.~Dai, Z.~Wang, D.~Wipf, \href{https://proceedings.mlr.press/v119/dai20c.html}{The usual suspects? reassessing blame for vae posterior collapse}, in: International Conference on Machine Learning, PMLR, 2020, pp. 2313--2322.

\bibitem{bowman2015generating}
S.~R. Bowman, L.~Vilnis, O.~Vinyals, A.~M. Dai, R.~Jozefowicz, S.~Bengio, \href{https://arxiv.org/abs/1511.06349}{Generating sentences from a continuous space}, arXiv preprint arXiv:1511.06349 (2015).

\bibitem{kingma2016improved}
D.~P. Kingma, T.~Salimans, R.~Jozefowicz, X.~Chen, I.~Sutskever, M.~Welling, \href{https://proceedings.neurips.cc/paper_files/paper/2016/hash/ddeebdeefdb7e7e7a697e1c3e3d8ef54-Abstract.html}{Improved variational inference with inverse autoregressive flow}, Advances in Neural Information Processing Systems 29 (2016).

\bibitem{sonderby2016ladder}
C.~K. S{\o}nderby, T.~Raiko, L.~Maal{\o}e, S.~K. S{\o}nderby, O.~Winther, \href{https://proceedings.neurips.cc/paper/2016/hash/6ae07dcb33ec3b7c814df797cbda0f87-Abstract.html}{Ladder variational autoencoders}, Advances in Neural Information Processing Systems 29 (2016).

\bibitem{huang2018improving}
C.-W. Huang, S.~Tan, A.~Lacoste, A.~C. Courville, \href{https://proceedings.neurips.cc/paper_files/paper/2018/hash/65b0df23fd2d449ae1e4b2d27151d73b-Abstract.html}{Improving explorability in variational inference with annealed variational objectives}, Advances in Neural Information Processing Systems 31 (2018).

\bibitem{higgins2017beta}
I.~Higgins, L.~Matthey, A.~Pal, C.~Burgess, X.~Glorot, M.~Botvinick, S.~Mohamed, A.~Lerchner, \href{https://openreview.net/forum?id=Sy2fzU9gl}{beta-vae: Learning basic visual concepts with a constrained variational framework}, in: International Conference on Learning Representations, 2017.


\bibitem{hoffman2016elbo}
M.~D. Hoffman, M.~J. Johnson, Elbo surgery: yet another way to carve up the variational evidence lower bound, in: Workshop in Advances in Approximate Bayesian Inference, NIPS, Vol.~1, 2016.

\bibitem{liu2024improving}
Y.~Liu, Z.~Yu, Z.~Liu, Z.~Yu, X.~Yang, X.~Li, Y.~Guo, Q.~Liu, G.~Wang, \href{https://www.sciencedirect.com/science/article/pii/S0950705124004520}{Improving disentanglement in variational auto-encoders via feature imbalance-informed dimension weighting}, Knowledge-Based Systems 296 (2024) 111818.

\bibitem{liu2023cloud}
Y.~Liu, Z.~Liu, S.~Li, Z.~Yu, Y.~Guo, Q.~Liu, G.~Wang, \href{https://www.sciencedirect.com/science/article/pii/S0031320323002303}{Cloud-vae: Variational autoencoder with concepts embedded}, Pattern Recognition 140 (2023) 109530.

\bibitem{gulrajani2016pixelvae}
I.~Gulrajani, K.~Kumar, F.~Ahmed, A.~A. Taiga, F.~Visin, D.~Vazquez, A.~Courville, \href{https://arxiv.org/abs/1611.05013}{Pixelvae: A latent variable model for natural images}, arXiv preprint arXiv:1611.05013 (2016).

\bibitem{ichikawa2025high}
Y.~Ichikawa, K.~Hukushima, \href{https://iopscience.iop.org/article/10.1088/1742-5468/adde3e/meta#jstatadde3es7}{High-dimensional asymptotics of vaes: threshold of posterior collapse and dataset-size dependence of rate-distortion curve}, Journal of Statistical Mechanics: Theory and Experiment 2025~(7) (2025) 073402.

\bibitem{cremer2018inference}
C.~Cremer, X.~Li, D.~Duvenaud, \href{https://proceedings.mlr.press/v80/cremer18a.html}{Inference suboptimality in variational autoencoders}, in: International Conference on Machine Learning, PMLR, 2018, pp. 1078--1086.

\bibitem{rezende2018taming}
D.~J. Rezende, F.~Viola, \href{https://arxiv.org/abs/1810.00597}{Taming vaes}, arXiv preprint arXiv:1810.00597 (2018).

\bibitem{tipping1999probabilistic}
M.~E. Tipping, C.~M. Bishop, \href{https://academic.oup.com/jrsssb/article-abstract/61/3/611/7083217}{Probabilistic principal component analysis}, Journal of the Royal Statistical Society Series B: Statistical Methodology 61~(3) (1999) 611--622.

\bibitem{kunin2019loss}
D.~Kunin, J.~Bloom, A.~Goeva, C.~Seed, \href{https://proceedings.mlr.press/v97/kunin19a.html}{Loss landscapes of regularized linear autoencoders}, in: International Conference on Machine Learning, PMLR, 2019, pp. 3560--3569.

\bibitem{Krizhevsky09learningmultiple}
A.~Krizhevsky, G.~Hinton, et~al., \href{https://www.cs.toronto.edu/~kriz/cifar.html}{Learning multiple layers of features from tiny images} (2009).

\bibitem{xiao2017fashion}
H.~Xiao, K.~Rasul, R.~Vollgraf, \href{https://arxiv.org/abs/1708.07747}{Fashion-mnist: a novel image dataset for benchmarking machine learning algorithms}, arXiv preprint arXiv:1708.07747 (2017).

\bibitem{kessy2018optimal}
A.~Kessy, A.~Lewin, K.~Strimmer, \href{https://www.tandfonline.com/doi/full/10.1080/00031305.2016.1277159}{Optimal whitening and decorrelation}, The American Statistician 72~(4) (2018) 309--314.

\bibitem{fai2024special}
L.~C. Fai, Special Functions in Physics and Engineering: A renewed approach with applications, IOP Publishing, 2024.

\bibitem{zhang2006schur}
F.~Zhang, The Schur complement and its applications, Vol.~4, Springer Science \& Business Media, 2006.

\end{thebibliography}

\end{document}